\documentclass[lettersize,journal]{IEEEtran}
\usepackage{amsmath,amsfonts}
\usepackage{algorithmic}
\usepackage{algorithm}
\usepackage{array}
\usepackage[caption=false,font=normalsize,labelfont=sf,textfont=sf]{subfig}
\usepackage{textcomp}
\usepackage{stfloats}
\usepackage{url}
\usepackage{verbatim}
\usepackage{graphicx}
\usepackage{cite}
\usepackage{multicol}
\usepackage{multirow}
\usepackage{booktabs}
\usepackage{orcidlink}
\usepackage{afterpage}
\usepackage{xcolor}
\usepackage{footnote}
\usepackage{tablefootnote}
\usepackage{arydshln}

\hypersetup{
    colorlinks=true,
    linkcolor=black,
    urlcolor=black,
    citecolor=black
}
\begin{document}

\title{Vision Language Models in Autonomous Driving: A Survey and Outlook}

\author{Xingcheng Zhou \orcidlink{0000-0003-1178-5221}, \and
Mingyu Liu \orcidlink{0000-0002-8752-7950}, \and
Ekim Yurtsever \orcidlink{0000-0002-3103-6052}, \IEEEmembership{Member, IEEE}, \and
Bare Luka Zagar \orcidlink{0000-0001-5026-3368}, \and \\
Walter Zimmer \orcidlink{0000-0003-4565-1272},  \and 
Hu Cao \orcidlink{0000-0001-8225-858X},  \and 
Alois C. Knoll \orcidlink{0000-0003-4840-076X}, \IEEEmembership{Fellow, IEEE} 

\thanks{This research is accomplished within the project ”AUTOtech.agil” (Grant Number 01IS22088U). We acknowledge the financial support for the project by the Federal Ministry of Education and Research of Germany (BMBF). (Corresponding author: Xingcheng Zhou)}
\thanks{X. Zhou, M. Liu, BL. Zagar, W. Zimmer, H. Cao, and AC. Knoll are with the Chair of Robotics, Artiﬁcial Intelligence and Real-Time Systems, Technical University of Munich, 85748 Garching bei München, Germany (E-mail: xingcheng.zhou@tum.de, mingyu.liu@tum.de, bare.luka.zagar@tum.de, walter.zimmer@tum.de, hu.cao@tum.de, knoll@in.tum.de)}
\thanks{E. Yurtsever is with the College of Engineering, Center for Automotive Research, The Ohio State University, Columbus, OH 43212, USA (E-mail: yurtsever.2@osu.edu)}}

\maketitle

\begin{abstract}
The applications of Vision-Language Models (VLMs) in the field of Autonomous Driving (AD) have attracted widespread attention due to their outstanding performance and the ability to leverage Large Language Models (LLMs). By incorporating language data, driving systems can gain a better understanding of real-world environments, thereby enhancing driving safety and efficiency. In this work, we present a comprehensive and systematic survey of the advances in vision language models in this domain, encompassing perception and understanding, navigation and planning, decision-making and control, end-to-end autonomous driving, and data generation. We introduce the mainstream VLM tasks in AD and the commonly utilized metrics. Additionally, we review current studies and applications in various areas and summarize the existing language-enhanced autonomous driving datasets thoroughly. Lastly, we discuss the benefits and challenges of VLMs in AD and provide researchers with the current research gaps and future trends. \href{https://github.com/ge25nab/Awesome-VLM-AD-ITS}{\color{red}{https://github.com/ge25nab/Awesome-VLM-AD-ITS}}

\end{abstract}

\begin{IEEEkeywords}
Vision Language Model, Large Language Model, Autonomous Driving, Intelligent Vehicle, Conditional Data Generation, Decision Making, Language-Guided Navigation, End-to-End Autonomous Driving.
\end{IEEEkeywords}

\section{Introduction}
\label{sec:intro}

    \IEEEPARstart{I}{ntelligent} mobility is important in modern civilization, driving economic growth, supporting urban development, and strengthening social connections. In recent years, the rapid evolution of deep learning and computing power has profoundly influenced this area, enhancing its efficiency and intelligence. Specifically, autonomous driving, as an integral part of this advancement, has developed rapidly and made many significant technological breakthroughs.

\begin{figure}[h!]
    \centering
    \includegraphics[width=0.4\textwidth]{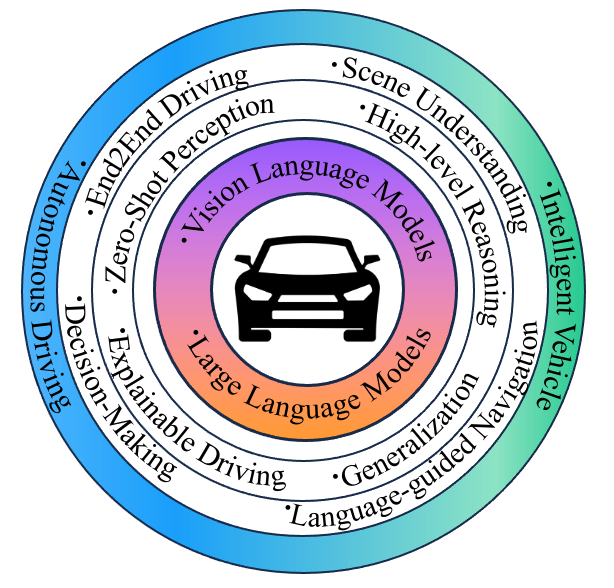}
        \caption{Vision-Language Models and Large Language Models offer advancements in traditional tasks and pave the way for innovative applications in AD.}
    \label{fig:figure1}
\end{figure}

Autonomous driving strives to enable vehicles to drive independently and intelligently. The current autonomous driving technologies, especially those related to perception and prediction, have tremendously benefited from advances in computer vision. For instance, perception modules, typically using Convolutional Neural Networks (CNNs) or Transformers \cite{transformerpaper}, process data from sensors like cameras or LiDAR to accurately identify and localize entities in their surroundings. However, despite these technological strides, the current autonomous driving solutions still heavily struggle in many dimensions, including handling complex and rapidly dynamic environments, explaining decisions, and following human instructions. They often fail to capture intricate details or understand context, leading to potential safety concerns and limiting the move toward more advanced autonomous driving.

\begin{figure*}[h!]
    \centering
    \includegraphics[width=0.9\textwidth]{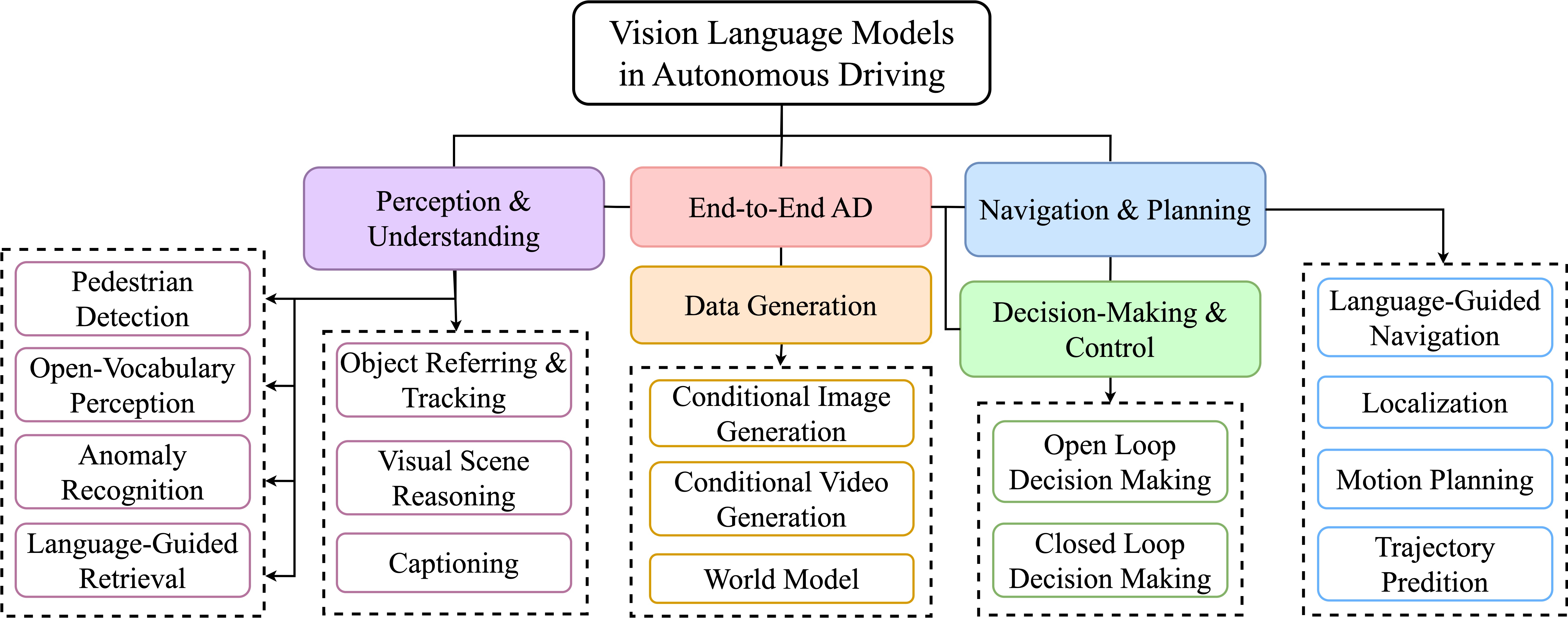}
        \caption{An Overview of the Taxonomy of Vision-Language Models Tasks and Applications in Autonomous Driving. This paper encompasses five major aspects of autonomous driving, i.e., Perception and understanding, Navigation and planning, decision-making and control, end-to-end autonomous driving, and data generation. We illustrate the main tasks and techniques inside the dashed rectangular boxes for each of these five dimensions. }
    \label{fig:taxonomy}
\end{figure*}

The emergence of LLMs \cite{LLMs1llama2,LLMs2LLama,LLMs3Bloom,LLMs4palm,LLMs5GPT3} and 
VLMs \cite{zhu2023minigpt,peng2023kosmos,gong2023multimodal,liu2023prismer,wu2023visual,zhang2023vision,bai2023qwenvl} 
provides potential solutions for the inherent limitations of current autonomous driving. These novel technologies synthesize linguistic and visual data, promising a future where vehicles and systems deeply understand their surroundings and natural languages. This indicates a new era of intelligent, efficient, and explainable transportation. Besides enhancing traditional tasks in AD, such as object detection or traffic prediction, emerging domains include zero-shot perception and language-guided navigation, as shown in Fig.~\ref{fig:figure1}. Given the surge in research applying language models to autonomous driving and intelligent systems, a systematic and comprehensive survey is of great importance to the research community. However, existing surveys \cite{zhang2023vision,End-to-endADSurvy,PromptinVLMsSurvey,ASurveyofLLMsZhao2023,ML4ITS:Asurvey} focus either on LLMs, VLMs, or AD individually. To the best of our knowledge, there is still no review or survey that systematically summarizes and discusses the applications of large VLMs in AD.

To this end, we present an extensive review of the current status of vision-language models in autonomous driving, highlighting the recent technological trends in the research community. We illustrate the taxonomy of this paper in Fig.~\ref{fig:taxonomy}. 
The main contributions of this work are summarized as follows:
\begin{itemize}
    \item We present a comprehensive survey on the large vision-language models in autonomous driving and categorize the existing studies according to VLM types and application domains.
    \item We consolidate the emergent mainstream vision language tasks and the corresponding common metrics in the area of autonomous driving. 
    \item We systematically summarize and analyze the existing classic and language-enhanced autonomous driving datasets. 
    \item We explore potential applications and technological advances of VLMs in autonomous driving.
    \item We provide an in-depth discussion on the benefits, challenges, and research gaps in this domain.
\end{itemize}

\section{Background} 
\label{background} 
This section gives a thorough overview of the related backgrounds, delving into the foundational concepts underlying these areas: Autonomous Driving (\ref{autonomous_driving}), Large Language Model (\ref{large_language_model}), and Vision-Language Model (\ref{vision_language_model}).
\subsection{Autonomous Driving} 
\label{autonomous_driving}

The \textit{Society of Automotive Engineers} (SAE) \cite{sae} introduced a classification system to describe and guide the development of AVs, which divides the driving automation level from Level 0 (No Automation) to Level 5 (Full Automation). As autonomy increases, human intervention reduces, while the requirements for the vehicle to understand its surroundings increase. As of now, the most advanced commercial vehicles are still at Level 2 or Level 3~\cite{meng2023configuration}, providing partial automation but still requiring driver supervision, and reaching Level 5 still has a long way to go. Recent study \cite{li2024survey} points out that techniques in intelligent transportation systems, including V2X cooperative perception \cite{zimmer2024tumtrafv2x}, and traffic flow forecasting \cite{Jiang_2022,ju2024cool,10184800} are also crucial components in the technical roadmap of achieving fully autonomous driving. Current mainstream autonomous driving systems can be classified into two types based on the differences in the pipeline: Classic modular paradigm and end-to-end approach. 

\textbf{Modular Autonomous Driving.} Modular autonomous driving system is a classic and widely adopted architecture in the domain\cite{yurtsever2020survey}. It divides the autonomous driving task into several standalone components: Perception, Prediction, Planning, and Control. Each module is developed separately and is responsible for specific functionality in the whole system. 

The perception component is the basis of the modular autonomous driving system. It collects, processes, comprehends, and interprets the vehicle's surrounding environmental information by leveraging onboard sensors. Depending on different technical road maps, the perception module utilizes one or multiple types of sensors, such as LiDAR, camera, radar, or event-based camera, to extract the location, size, velocity, and other details of traffic participants of interest, i.e., vehicles, pedestrians, obstacles, and lane markings. Perception module typically involves various tasks and technologies from computer vision, encompassing  2D and 3D object detection~\cite{lang2019pointpillars,erccelik20223d,zimmer2023real,detr,li2022bevformer}, semantic segmentation \cite{3dsegement}, multiple object tracking \cite{motsuervey}, and sensor fusion \cite{sensorfusion}, among others. The prediction module is built upon the real-time environmental perceptive information from the perception module. By analyzing the historical trajectories and current status of the road users of interest, it aims to predict their short-term and long-term trajectories and behaviors \cite{mangalam2021goals, gilles2021home, gilles2022gohome}. This includes predicting the future behavior of vulnerable traffic participants, such as pedestrians and cyclists, as well as the driving strategies of other vehicles. The prediction module plays a vital role in ensuring the safety and stability of driving by forecasting future traffic conditions. 

The planning module is responsible for generating the optimal path from the current location to the target destination based on the vehicle's state and environmental information. Typically, this process can be divided into global planning and local planning. Global planning focuses on the optimal route planning from the starting point to the destination, where methods like the A* \cite{a*paper} or Dijkstra algorithms are used to search on a map. On the other hand, local planning involves real-time adjustments based on the specific current situation of the vehicle, with a greater emphasis on immediacy and reliability. Common methods for local planning include RPP\cite{RPP}, RRT\cite{RRT}, RRT*\cite{rrt*}, etc. Additionally, many deep learning-based planners \cite{Motion_Planning_Teng_2023,MP1,MP2,MP3} have emerged as powerful alternatives to traditional methods in recent years. The control module takes charge of the executing of the trajectory and path given by the planning module. It takes into account the vehicle's inherent characteristics, such as the vehicle dynamics model, as well as external environmental conditions, and adjusts a range of control signals, such as acceleration and steering, to control the vehicle in a stable and robust manner.

\textbf{End-to-End Autonomous Driving.} Compared with the modular autonomous driving paradigm, end-to-end autonomous driving \cite{bojarski2016end, sadat2020perceive, hu2023planning,jiang2023vad} endeavors to integrate separate components into a unified system and optimizes the whole system in a differentiable way. It treats autonomous driving as a singular learning task, taking raw data inputs directly from various sensors and outputting the corresponding control signals. The primary goal is to optimize the final planning performance, ensuring that every decision made by the system enhances the overall driving experience. By focusing on the main objective, end-to-end AD promotes an integrated and efficient decision-making process, ultimately contributing to safer and more reliable autonomous driving. The advantages of end-to-end autonomous driving are evident in its straightforward and compact architecture, clearer and more direct objectives, reduced reliance on rule-based design, and performance scalability with more training resources\cite{End-to-End-survey2,End-to-endADSurvy}.

However, both modular and end-to-end autonomous driving schemes still face serious challenges, such as generalization, interpretability, causal confusion, and robustness. Although many studies have attempted to address these issues using various methods, how to construct a safe, stable, and interpretable AD system remains an open topic. The excellent benefits of Large Vision Language Models in scene understanding, reasoning, zero-shot recognition, and desirable interpretability provide a new solution and research direction for the community to overcome these challenges.

\subsection{Large Language Models}
\label{large_language_model}

Natural language is the primary means by which humans communicate and convey information. Natural language models, designed to comprehend and process natural language, have evolved through years of development. The introduction of Transformer architecture \cite{transformerpaper} brought about a disruptive revolution in the Natural Language Processing (NLP) field due to its highly parallelized data processing mechanism and powerful performance. BERT\cite{devlin2019bert}, as another milestone of natural language models, proposed an approach to pre-train the model over vast amounts of unlabeled corpora data, followed by fine-tuning on specific tasks. It significantly enhances the baseline performance across various NLP tasks. This approach of learning context-aware feature representations from large-scale unlabeled data has laid the groundwork for the emergence of large models. 

Large Language Models (LLMs) typically refer to language models with a massive number of parameters, often in the order of a billion or more. A recent study \cite{kaplan2020scalinglaw} shows that the performance of language models depends on factors such as the number of model parameters, the size of the dataset, and the amount of training computation, and formulates the scaling law. The most notable characteristic of LLMs is the exhibition of emergent abilities, such as the capacity for few-shot or zero-shot transfer learning across numerous downstream tasks, strong multi-step reasoning capabilities, and the ability to follow instructions, which are normally not present in smaller models. ChatGPT, specifically GPT-3.5 \cite{LLMs5GPT3} and GPT-4 \cite{VLM1OpenAI2023GPT4TR}, serves as a milestone in the development of LLMs. Since its release, GPT-3.5 has consistently drawn attention due to its exceptional performance. An increasing number of researchers are beginning to explore and harness LLMs' powerful linguistic understanding, interpretation, analysis, and reasoning capabilities to solve previously hard or impossible problems. In addition, a number of open-source LLMs, i.e., Llama2 \cite{LLMs1llama2}, Qwen \cite{bai2023qwen}, Phi\cite{phi-1.5}, etc., also gain considerable attention from academia and industry, and some of them even achieve comparable performance to ChatGPT in specific tasks.

\begin{figure}[t]
    \centering
   \includegraphics[width=0.46\textwidth]{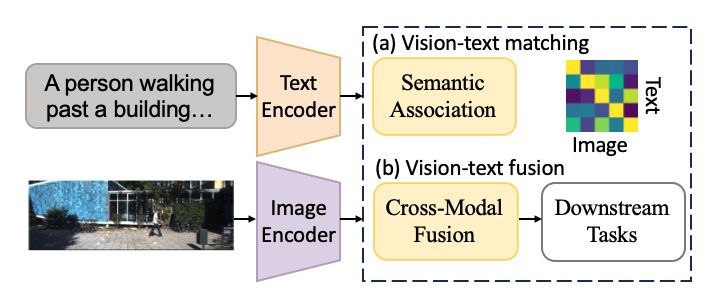}
    \caption{Two inter-modality connection approaches of Vision-Language Models in Autonomous Driving: \textbf{(a)} Vision-text matching. We demonstrate the semantic similarity matching in the top-right of this figure. \textbf{(b)} Vision-text fusion. The fused features can be used for downstream tasks. The figure is from the KITTI \cite{geiger2013vision} dataset.}
    \label{fig:2typeVLMs}
\end{figure}

\subsection{Vision-Language Models}
\label{vision_language_model}
Vision-Language Models (VLMs) bridge the capabilities of Natural Language Processing (NLP) and Computer Vision (CV), breaking down the boundaries between text and visual information to connect multimodal data. VLMs can comprehend the complex relationship between visual content and natural languages by learning cross-modality data. Recently, with the rise of LLMs, there is also an increasing focus on exploring how to effectively incorporate visual modules into LLMs to perform multimodal tasks.

\begin{figure*}[t]
    \centering
    \includegraphics[width=1\textwidth]{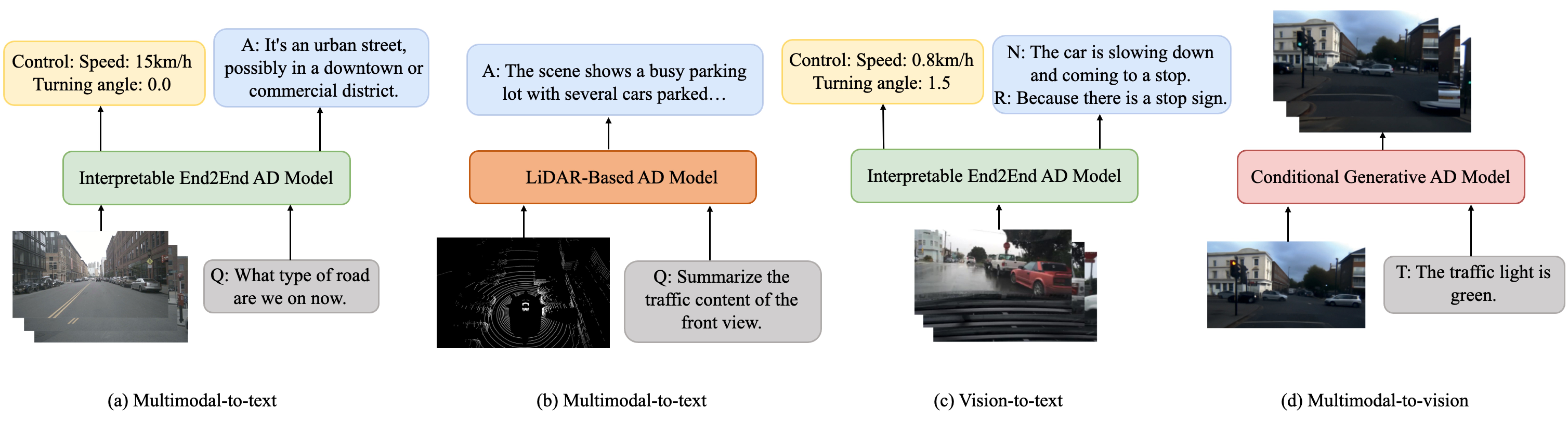}\label{fig:v2t}
    \caption{Overview of mainstream Vision-Language Models in Autonomous Driving. \textbf{(a)} Multimodal-to-text models take text and image or video as input and generate text, as in \cite{xu2023drivegpt4}. \textbf{(b)} Multimodal-to-text models take text and point clouds as input and generate text, as in \cite{LiDAR-LLM}. \textbf{(c)} Vision-to-text models accept video or image as input and produce text as output, e.g. GAIA-1 \cite{Hu2023GAIA-1}. \textbf{(d)} Multimodal-to-vision models take image and text as input and output image or video, depicted with \cite{jin2023adapt}.}
    \label{fig:3typeVLMs}
\end{figure*}

Mainstream Vision-Language Models in AD can be broadly categorized into Multimodal-to-Text (M2T) \cite{xu2023drivegpt4}\cite{qian2023nuscenesQA,fu2023drivelikeahuman,dewangan2023talk2bev}, Multimodal-to-Vision (M2V) \cite{Hu2023GAIA-1}\cite{wang2023drivedreamer}, and  Vision-to-Text (V2T) \cite{jin2023adapt}\cite{Wei2020NIC} based on input-output modality types, as shown in Fig.~\ref{fig:3typeVLMs}. M2T type typically takes image-text or video-text as input and produces text as output. LiDAR, as another commonly used sensor in autonomous driving, can also formulate the M2T-type vision language model. Correspondingly, the M2V-type models accept image-text as input and generate image or video as output, while V2T-type models take image or video as input and generate texts as output. As illustrated in Fig.~\ref{fig:2typeVLMs}, according to the inter-modality information connection approaches, VLMs employed in AD are divided into Vision-Text-Fusion (VTF) \cite{xu2023drivegpt4}\cite{jin2023adapt}\cite{qian2023nuscenesQA}\cite{Cheng2023MSSG, talktothevehicle, CovertNet} and Vision-Text-Matching (VTM) \cite{Wei2020NIC}\cite{liu2023vlpd, liu2023umpd, wu2023referrkitti, wu2023nuprompt, peng2023openscene, chen2023clip2scene, najibi2023upvl, romero2023zelda, jain2023groundthennavigate, omama2023alt}. VTF-type vision language models employ various fusion methods to effectively integrate vision embedding and language embedding and jointly optimize the feature representation that performs better for the target task. In contrast, VTM-type vision language models, including image-text matching \cite{CLIP,cherti2022openclip} and video-text 
 matching \cite{xu2021videoclip}, learn a joint representation space by forcing vision-text pairs semantically close to each other and unpaired instances distant to each other, achieving cross-modal semantic alignment and enabling cross-modal semantic propagation. CLIP \cite{CLIP}, a milestone image-text matching work in VLMs, captures image feature representations associated with language and achieves zero-shot transfer ability by training on a vast number of image-text pairs through contrastive learning.

\section{Vision-Language Tasks in Autonomous Driving}
\label{vlm_task_ad}
As researchers become increasingly aware of the significance of language integration in autonomous driving, more attention is being directed towards studying the vision language tasks in autonomous driving. In this section, we summarize the existing primary vision language tasks in autonomous driving, i.e., Object Referring and Tracking in (\ref{Object_referring_and_tracking}), Open-Vocabulary Traffic Environment Perception (\ref{Open-Vocabulary Traffic Environment Perception}), Traffic Scene Understanding (\ref{Traffic Scene Understanding task}), Language-Guided Navigation (\ref{Language-Guided Navigation}), and Conditional Autonomous Driving Data Generation (\ref{data_generation_tasks}). We introduce the problem definitions, clarify the difference with non-VLM tasks, and formulate the main evaluation metrics.

\subsection{Object Referring and Tracking}
\label{Object_referring_and_tracking}

\textbf{Single or Multiple Object Referring.} 
Object Referring (OR), also known as object referral or object grounding in some papers, aims to localize the specified object in 2D images or 3D spaces using a given natural language expression. When a language prompt description corresponds to a single object in the scene, it is termed single object referring (SOR). When the language prompt description refers to multiple objects simultaneously, it is termed multiple object referring (MOR). OR can be considered a type of language-conditional object detection. In contrast to the classic object detection task, which aims at localizing and classifying some specified categories, the OR task specifies the object or multiple objects to be detected via language prompt. Hence, OR can be considered a generalized object detection task. \\

\begin{figure}[htb]
    \centering
   \includegraphics[width=0.48\textwidth]{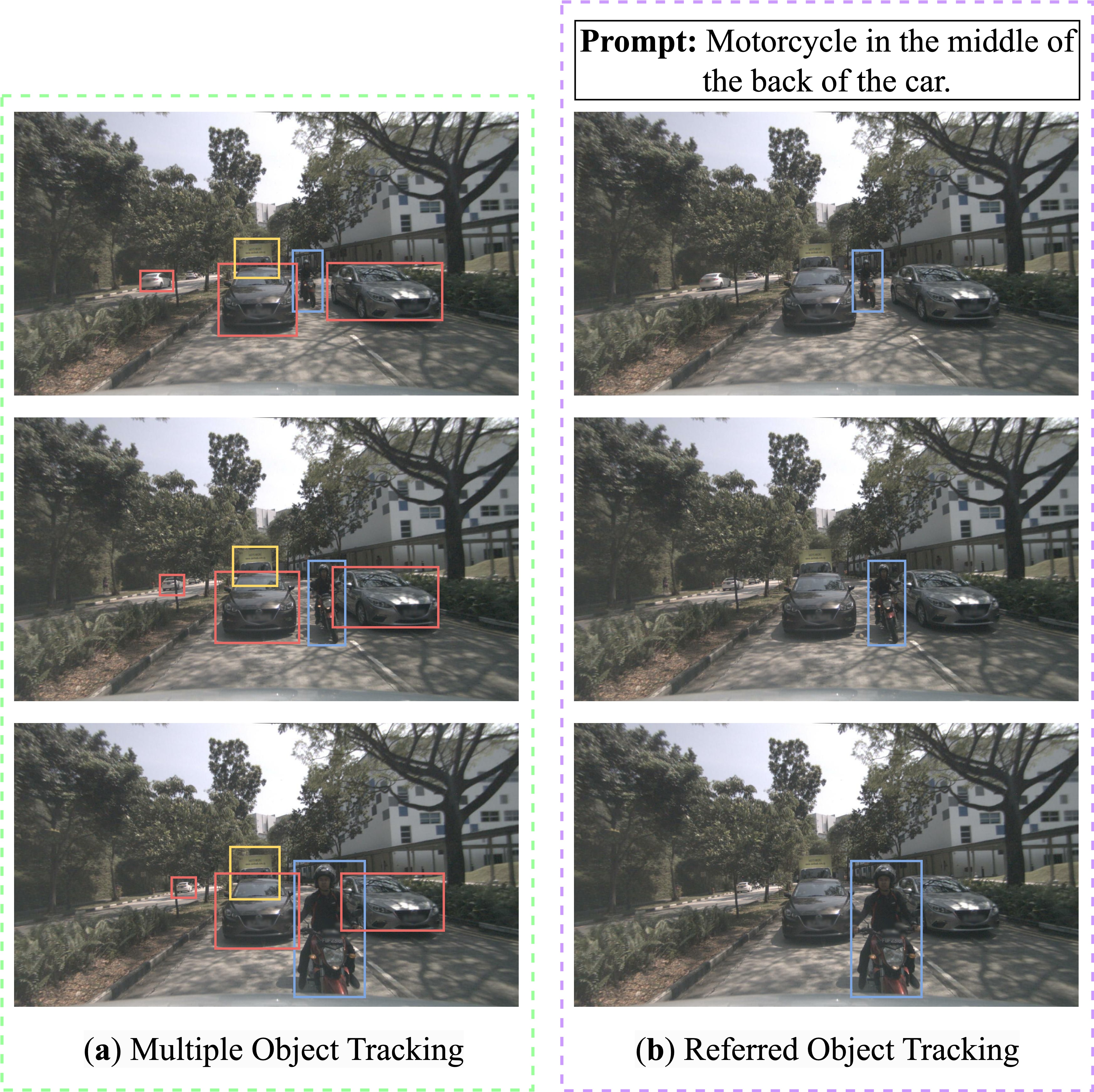}
    \caption{Example comparison between classic multiple object tracking task (left) and referred object tracking task (right). The image sequences are extracted from the rear camera of the nuScenes \cite{caesar2020nuscenes} Dataset.}
    \label{fig:odvsor}
\end{figure}

\textbf{Referred Object Tracking.}
In addition to the object-referring task, some recent works have introduced the new object-referring and tracking (OR-T) task in autonomous driving. Compared to the object-referring task, the OR-T task takes language expressions as semantic cues to track one or more objects across consecutive frames. OR-T aims to evaluate the tracking ability of specified single or multiple objects by language prompting in the video input. As shown in Fig. \ref{fig:odvsor}, compared with traditional tracking tasks, OR-T focuses more on the cross-frame, referring to consistency and robustness. Hence, the OR-T task can be considered a generalized traditional object-tracking task that tracks specific objects specified by language prompts. When all objects are designated as tracking targets, the OR-T  can be seen as equivalent to the traditional object-tracking task. \\

\textbf{Evaluation Metrics.}
Since OR and OR-T tasks in autonomous driving are generally designed and tested based on mainstream autonomous driving datasets, such as nuScenes \cite{caesar2020nuscenes} and KITTI \cite{geiger2013vision}, they also adopt the same evaluation metrics as used in the dataset's benchmarks. The Object Referring task uses the mean Average Precision (mAP) to measure the model's accuracy in localizing specified objects. For the Multiple OR-T task, metrics used in multiple object tracking (MOT) tasks, such as Higher Order Tracking Accuracy (HOTA), Multi-Object Tracking Accuracy (MOTA) in Eq. \ref{AMOTA}, Multi-Object Tracking Precision (MOTP) in Eq. \ref{AMOTP}, and Identity Switches (IDS) are adopted as evaluation standards, where FN, FP, IDS represents false negative, false positive, identity switches, and $d(o_i, h_i)$ represents the i-th distance between tracked and ground truth object respectively.

\begin{equation} \label{AMOTA}
\textbf{MOTA} = 1 - \frac{FN_t + FP_t + IDS_t}{GT}
\end{equation}

\begin{equation} \label{AMOTP}
\textbf{MOTP} = \frac{\sum_{i=1}^{M} d(o_i, h_i)}{M}
\end{equation}

\subsection{Open-Vocabulary Traffic Environment Perception}
\label{Open-Vocabulary Traffic Environment Perception}
\textbf{Open-Vocabulary 3D Object Detection.} Open-Vocabulary 3D Object Detection (OV-3DOD), also known as zero-shot 3D Object Detection, aims to improve the performance of detecting objects with novel categories that are not included in the training process. In contrast to open-set 3D object detection, which only differentiates known classes from unseen ones, OV-3DOD extends all novel categories of interest as a vocabulary of text prompts and assigns a specific semantic category in the vocabulary to each novel 3D bounding box. The primary challenges of OV-3DOD revolve around two main questions: how to generate novel 3D bounding boxes and determine the classes of these boxes. Regarding the first question, since generating novel 3D bounding boxes based on 2D images is relatively difficult, most current studies leverage point cloud data provided by LiDAR sensors for new 3D bounding box proposals. Addressing the latter issue, existing works attempt to align the semantic features of the 3D bounding boxes to the feature domain of VTM-based VLMs to leverage their strong zero-shot inference abilities. \\ 

\textbf{Open-Vocabulary 3D Semantic Segmentation.} Open-Vocabulary 3D Semantic Segmentation (OV-3DSS), sometimes also formulated as zero-shot learning of 3D point clouds, is designed to segment 3D point clouds or 3D meshes into semantically meaningful regions where the classes do not exist in the set of predefined classes during training. Like OV-3DOD, OV-3DSS seeks to discover the novel points or meshes from novel objects and classify them into unseen categories in a zero-shot manner. The mainstream studies also focus on aligning point-level or mesh-level features to the corresponding text features leveraging VTM-based large VLMs. \\

\textbf{Evaluation Metrics.} The primary difference between OV-3DOD and classic 3D Object Detection is whether the object categories of interest are open-vocabulary or closed-set. During evaluation, the same metrics are applied, hence OV-3DOD also utilizes the 3D mAP for accessing the model's performance. Similarly, the OV-3DSS task employs the same metrics as classic 3D Semantic Segmentation, using mIOU and mean Accuracy (mAcc) for evaluation.

\subsection{Traffic Scene Understanding}
\label{Traffic Scene Understanding task}

\begin{figure}[htb]
    \centering
   \includegraphics[width=0.49\textwidth]{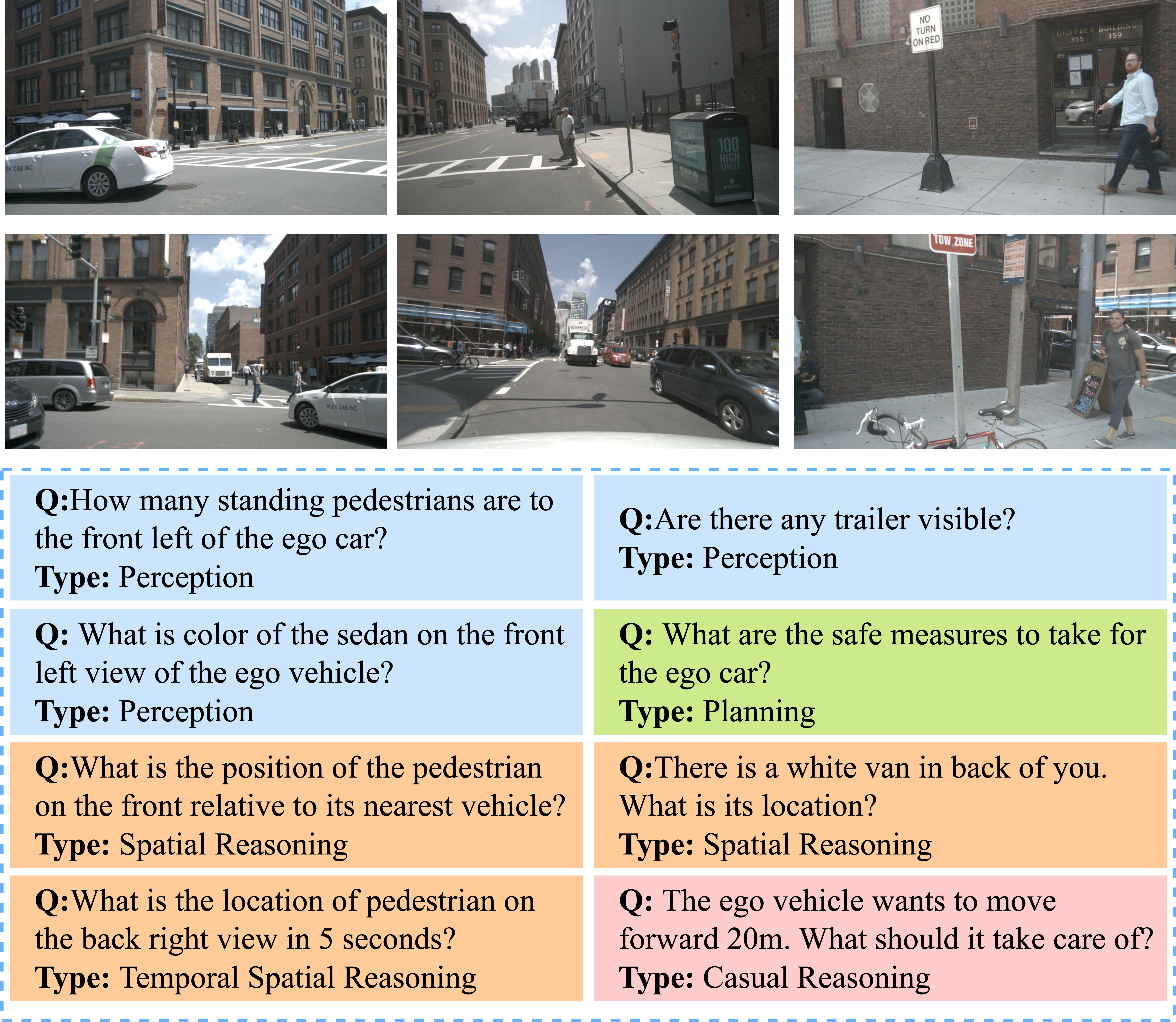}
    \caption{Example of different types of questions commonly raised in VQA tasks in autonomous driving. Depending on the settings, the visual input can be multiple images as shown above, or single cameras, point clouds, and video streams. The multi-view images and question examples are selected from \cite{caesar2020nuscenes,Reason2Drive,dewangan2023talk2bev,LiDAR-LLM}. The taxonomy of question types can be slightly different in each dataset.}
    \label{fig:exampquestions}
\end{figure}

\textbf{Visual Question Answering.} Traffic Scene Visual Question Answering (VQA) involves answering questions based on images or video input, which is a challenging task requiring a high-level understanding of the traffic scene and proposed questions. It offers a potential avenue to address the challenges of interpretability and trustworthiness in traditional autonomous driving systems. Questions of interest in autonomous driving can be generally categorized into perception, planning, spatial reasoning, temporal reasoning, and causal reasoning. Perception questions are designed to recognize and identify the traffic participants with basic tasks, such as appearance description, presence, counting, status, etc. Planning questions serve to make sound decisions and actions to accomplish the goals based on current traffic conditions. Spatial reasoning questions aim to ascertain the absolute and relative positions of objects of interest in 3D space, including their proximities and distances. Temporal reasoning questions involve inferring future or past behaviors and trajectories of objects based on existing video data. Causal reasoning questions are typically more complex, analyzing the reasons behind occurrences and generating anticipated outcomes through logical actions. This sometimes requires leveraging common sense. Figure \ref{fig:exampquestions} illustrates some question examples covering these types. \\

\textbf{Captioning.} Captioning is another common task for traffic scene understanding in autonomous driving. For a given scene, such as an image, a point cloud, or a specified object within it, the task involves generating a textual description of its content. Unlike the VQA task, which requires questions as input, captioning focuses more on specific tasks like scene description, importance ranking, action explanation, etc. Therefore, captioning can be considered a special case of VQA with a fixed question, often employing the same evaluation metrics as open-ended VQA.

\textbf{Evaluation Metrics.} In multiple-choice VQA tasks, the correct answer appears among the options in a multiple-choice format. In this case, Top-N Accuracy can be used to evaluate whether the model makes the correct prediction. As described by Eq. \ref{Topkacc}, the answer can be considered a successful prediction if the correct answer is among the top n choices ranked by the model's predicted probabilities. The accuracy is then calculated by dividing the number of successfully predicted questions by the total number of questions. The choice of N typically depends on the number of choices and the type of questions. NuScenes-QA\cite{qian2023nuscenesQA} and Talk2BEV \cite{dewangan2023talk2bev} adopt Top-1 accuracy as the evaluation metric.
\begin{equation} \label{Topkacc}
\textbf{Top-N Accuracy} = \frac{\# \text{Correct Top-N Predictions}}{\# \text{Total Questions}}
\end{equation}

In the setting of VQA with open-ended answering format and captioning, some common metrics are used to evaluate the relevance and correctness between the predicted answers and ground truth, such as BLEU\cite{BLEU}, METEOR\cite{meteor}, ROUGE\cite{ROUGE}, CIDEr\cite{cider}, etc. Bilingual Evaluation Understudy (BLEU) score, in Eq. \ref{BLEU}, multiplies the weighted average of n-gram precision by the Brevity Penalty (BP). BP acts as a penalty factor for the length of sentences, encouraging the generated answer's length to closely match that of the ground truth answer. The weighted average of N-Gram measures the similarity of key vocabulary between the prediction and GT, and N is used to refer to its several variants BLEU-N. 

\begin{equation} \label{BLEU}
\textbf{BLEU} = \text{BP} \cdot \exp\left(\sum_{n=1}^{N} w_n \log p_n\right)
\end{equation}

Each of these metrics has its focus and limitations. Therefore, current studies usually utilize them in combination to provide a more comprehensive evaluation of the VQA system. To mitigate the semantic meaning deficiency of these metrics, Reason2Drive \cite{Reason2Drive} proposes an evaluation protocol to measure the performance of the reasoning chain. Besides, some language models or LLMs can also be used as a metric for accessing the similarity between model predictions and reference answers via proper similarity calculation or prompts, such as BERTScore \cite{Zhang*2020BERTScore:}, GPT-3.5 Score\cite{Reason2Drive}, LLaMA score \cite{LLMs1llama2}, etc.

\subsection{Language-Guided Navigation}
\label{Language-Guided Navigation}

\textbf{Task Description.}
In autonomous driving, language-guided navigation typically refers to the task where vehicles make reasoned plans and reach specified locations based on language instructions. Due to the complexity, variability, and uncertainty of outdoor scenarios, this task is more challenging than the Language-guided navigation tasks in indoor settings. It often involves language-guided localization and path planning, where understanding the target location specified in the instructions is a prerequisite for successful navigation. Therefore, existing work often leverages the advantages of VTM-type VLMs to find the corresponding target location among landmarks in maps. \\

\textbf{Evaluation Metrics.}
A language-guided navigation system usually concerns the following aspects during the evaluation process: the ability to successfully complete the navigation task, the accuracy of target localization based on the given language prompt, and the deviation between the target location and its final arrival. Task Completion (TC) is a metric used to measure the success rate of completing the navigation task, where reaching the target location or its neighboring node in the environment can be considered successful navigation. Regarding measuring localization accuracy, some studies \cite{omama2023alt} utilize Recall@K as the metric and calculate the similarity between the K nearest landmarks of the ground truth position and the predictions. To assess the deviation between the final reached location and the target location, metrics like Shortest-path Distance (SPD) or Absolute Position Error (APE) can be used to measure the bias. Regarding the planning performance during the navigation process, some works\cite{jain2023groundthennavigate} employ metrics such as the Frechet Distance (FD) or normalized Dynamic Time Warping (nDTW) for evaluation.

\subsection{Conditional Autonomous Driving Data Generation}
\label{data_generation_tasks}

\textbf{Task Description.}
Conditional autonomous driving data generation, also known as controllable autonomous driving data generation, allows for the targeted acquisition of photorealistic synthetic data. This enhances the diversity of training data, enabling purposeful augmentation of the database for autonomous driving systems. Models for controllable data generation typically incorporate single or multiple conditions, such as text prompts, Bird's Eye View (BEV) masks or sketches, specific action states, bounding box positions, or HD maps. The data generated can be classified into single-view or multiple-view scenarios, encompassing image generation and video generation.

\textbf{Evaluation Metrics.}
Conditional autonomous driving data generation employs the same evaluation metrics as text-to-image and text-to-video tasks. Two commonly used metrics are Fréchet Inception Distance (FID) \cite{FID}, also known as Wasserstein-2 distance, and Fréchet Video Distance (FVD)\cite{FVD}. As shown in Eq. \ref{FID}, $(\mu_x,\Sigma_x)$ and $(\mu_g,\Sigma_g)$ denote the Gaussian distribution of the image features extracted from the generated synthetic image and ground truth image, respectively. The input in FVD is replaced with the feature distribution obtained from a pre-trained video backbone. Lower values of FID and FVD indicate smaller distribution differences between the generated and real data and represent higher quality of the images or videos. Table \ref{Data Generation Models Table} summarizes the performance of recent studies in conditional autonomous driving data generation with FID and FVD metrics.
\begin{equation} \label{FID}
\textbf{FID} = ||\mu_x - \mu_g||^2 + \text{Tr}(\Sigma_x + \Sigma_g - 2(\Sigma_x\Sigma_g)^{1/2})
\end{equation}

In addition to FID and FVD, BEVControl \cite{BEVcontrol} also introduces CLIP score as an extra metric for evaluating multi-view image generation. The CLIP score measures the visual consistency of generated multiple street views. For the work based on the nuScenes dataset, metrics such as mAP, NDS, and mAOE can also be used to analyze the effectiveness of foreground object conditions. These metrics more explicitly reflect the model's compliance with conditions in the context of autonomous driving scenario-based condition data generation tasks.

\section{Mainstream Methods and Techniques}
\label{vlm_ad}

An increasing number of initiatives are endeavoring to implement VLMs across various aspects of AD. This section summarizes the methods and techniques addressing the mainstream tasks mentioned in Section \ref{vlm_task_ad}. We present the existing studies of VLMs within AD consisting of Perception and Understanding (\ref{perception_understanding}), Navigation and Planning (\ref{naivigation_palnning}), Decision-Making and Control (\ref{Decision-Making and Control}), End-to-End Autonomous Driving (\ref{End-to-End Autonomous Driving}), and data generation (\ref{data_generation}). The summarized current methods are shown in Tab.~\ref{vlms_ad}.

\subsection{Perception and Understanding}
\label{perception_understanding}

\begin{figure*}[h]
    \centering
    \includegraphics[width=\textwidth]{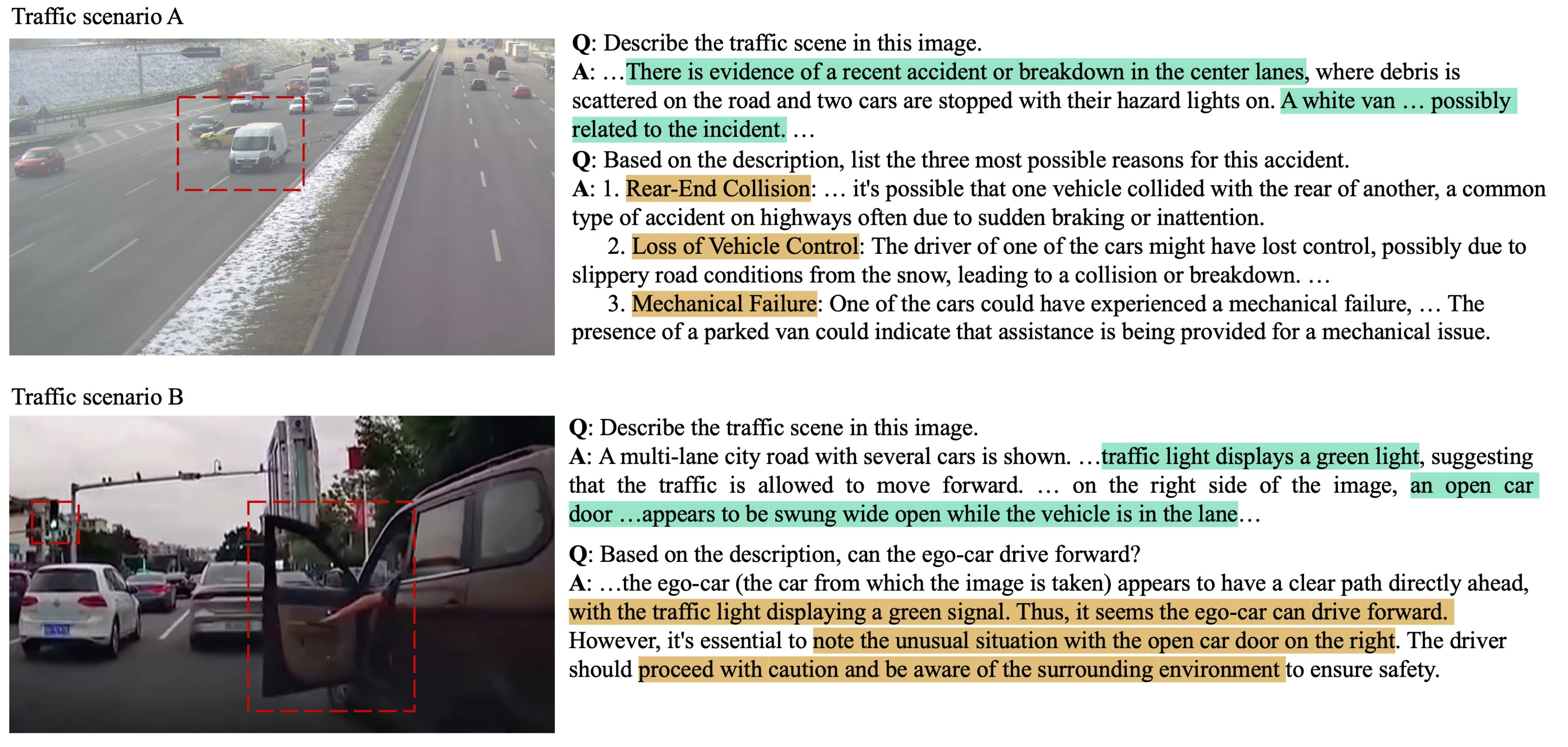}
    \caption{Evaluation of GPT-4V \cite{GPT4V} on traffic anomaly scene interpretability (green) and reasoning (yellow). The traffic scenario A is sourced from the A9 dataset~\cite{cress2022a9}. The traffic accident is highlighted in the red rectangle dotted box at the left-top of this figure. In the second row, we demonstrate traffic scenario B from \href{https://www.youtube.com/watch?v=CR5x89hZIYg}{YouTube}, where a car's door is open while the car is in the middle of the lane. The red dotted boxes show the opened door and the traffic light.}
    \label{fig:vlm_test}
\end{figure*}

In the perception module of autonomous driving, large Vision Language Models, especially those pre-trained on large-scale datasets with image-text-matching approach, like \cite{CLIP,zhang2024longclip}, have facilitated numerous new studies. These studies \cite{liu2023umpd,peng2023openscene} leverage the substantial prior knowledge of pre-trained large VLMs to boost the performance of perception and understanding and further introduce many novel tasks in the field. \\

\noindent \textbf{Object Referring.} As formulated in Section \ref{Object_referring_and_tracking}, compared to traditional perception tasks in AD, such as object detection, tracking, and semantic segmentation, the introduction of language enables the model to attain a more finely-grained and comprehensively unconstrained capability to understand the surrounding environment. Object Referring localizes the described objects using boxes or masks based on the language queries. MSSG \cite{Cheng2023MSSG} proposes a multi-modal 3D single object referring (SOR) task in autonomous driving scenarios. It trains a multi-modal single-shot grounding model by fusing the image, LiDAR, and language features under Bird Eye View (BEV) and predicts the targeted region directly from the detector without any post-processing. TransRMOT \cite{wu2023referrkitti} extends from the SOR task to Multiple-Object-Referring and Tracking (MOR-T), and constructs Refer-KITTI benchmark based on the KITTI dataset. Given a language query, TransRMOT can detect and track an arbitrary number of referent objects in videos. Similarly, PromptTrack \cite{wu2023nuprompt} proposes a language prompt set for the nuScenes dataset and constructs the NuPrompt benchmark. Compared to ReferKITTI, NuPrompt inherits nuScenes' multi-view attributes, making it applicable for multi-view MOR-T tasks. \\

\noindent \textbf{Open-Vocabulary 3D Object Detection and Semantic Segmentation.} According to the definition in Section \ref{Open-Vocabulary Traffic Environment Perception}, Open-Vocabulary Perception is a challenging task aiming at identifying unseen objects with specific categories. Due to the potent zero-shot transfer and cross-modal mapping capabilities of large VLMs, object detection and semantic segmentation are endowed with the ability to perform open-vocabulary detection and segmentation on unseen samples. OpenScene \cite{peng2023openscene} utilizes 2D-3D projective correspondence to enforce consistency between 3D point cloud embeddings and the corresponding fused 2D image features. Essentially, it aligns the 3D point cloud representation with CLIP's image-text representation to acquire zero-shot understanding capabilities for dense 3D point features. OpenScene is primarily evaluated in indoor scenarios but also demonstrates satisfactory open vocabulary 3D semantic segmentation (OV-3DSS) capabilities on nuScenes. Similarly, CLIP2Scene \cite{chen2023clip2scene} explores how to utilize CLIP to assist with 3D scene understanding in autonomous driving. By seeking connections in modality between pixel-text mapping and pixel-point mapping, CLIP2Scene constructs point-text pairs and pixel-point-text pairs for the training of contrastive learning, respectively. The objectiveness is also to ensure that the 3D point feature and its corresponding language achieve semantic consistency, thereby facilitating OV-3DSS. Experiments indicate that using CLIP2Scene as pre-training greatly outperforms other self-supervised methods. 

UP-VL framework \cite{najibi2023upvl} first presents an unsupervised multi-modal auto-labeling pipeline to generate point-level features and object-level bounding boxes and tracklets for open-vocabulary class-agnostic 3D detector supervision, which is further utilized for proposing 3D bounding box at inference time. Together with the assigned semantic labels through similarity matching, UP-VL framework realizes unsupervised open-vocabulary 3D detection and tracking (OV-3DOD-T) of static and moving traffic participants in AD scenarios.\\

\begin{table*}[!ht]
\centering
\caption{Overview of LLMs and VLMs in Autonomous Driving \\
\textbf{Tasks}: \textbf{PD}: Pedestrian Detection \textbf{SOR}: Single-Object Referring \textbf{MOR-T}: Multi-Object Referring and Tracking \textbf{OV-3DSS}: Open-Vocabulary 3D Semantic Segmentation \textbf{OV-3DOD-T}: Open-Vocabulary 3D Object Detection and Tracking \textbf{LGR}: Language-guided retrieval \textbf{IC}: Image Captioning \textbf{AR}: Anomaly Recognition  \textbf{VQA}: Visual Question Answering \textbf{3D-G-C}: 3D Grounding and Captioning \textbf{LGN}: Language-guided Navigation \textbf{MP}: Motion Planning \textbf{TP}: Trajectory Prediction \textbf{OL-DM}: Open-loop Decision Making \textbf{CL-DM}: Closed-loop Decision Making \textbf{OL-C}: Open-loop Control \textbf{OL-P}: Open-loop Planning \textbf{CVG}: Conditional Video Generation \textbf{CIG}: Conditional Image Generation  \\ 
\textbf{Type}: \textbf{V2T}: Vision-to-Text \textbf{VTM}: Vision-Text Matching \textbf{VTF}: Vision-Text Fusion \textbf{M2T}: Multimodal to Text \textbf{LLM}: Large Language Model
}
\resizebox{\textwidth}{!}{
\begin{tabular}{lcccl}
\toprule
\textbf{Method} & \textbf{Year} & \textbf{Tasks}& \textbf{Type} & \multicolumn{1}{l}{\textbf{Contribution}} \\ 
\midrule
\multicolumn{5}{c}{\textbf{Perception and Understanding}} \\ 
\midrule

VLPD \cite{liu2023vlpd}  &  2023    &   PD      &  VTM         &   Propose a vision-language extra-annotation-free method for pedestrian detection.           \\

UMPD \cite{liu2023umpd}   &  2023    &    PD  &  VTM       & Propose a fully unsupervised multi-view pedestrian detector without manual annotations.           \\

MSSG \cite{Cheng2023MSSG} &  2023 & SOR & VTF & Incorporate natural language with
3D LiDAR point cloud and images for LiDAR grounding task.  \\

TransRMOT \cite{wu2023referrkitti}    &   2023   &    MOR-T     &  VTM       &  Propose Referring  Multi-Object Tracking task, dataset, and benchmark based on KITTI.       \\ 

PromptTrack \cite{wu2023nuprompt}   &  2023    &     MOR-T          &   VTM      &  Propose Multi-view 3D Referring Multi-Object Tracking dataset based on nuScenes.            \\

  
OpenScene \cite{peng2023openscene}   &   2023   &     OV-3DSS        & VTM      &   Propose a zero-shot method for extracting 3D dense features from open vocabulary embedding space. \\ 

CLIP2Scene \cite{chen2023clip2scene}  & 2023 & OV-3DSS & VTM & Distill CLIP knowledge to a 3D network for 3D scene understanding. \\

UP-VL \cite{najibi2023upvl} & 2023 & OV-3DOD-T & VTM &  Introduce semantic-aware unsupervised detection for objects in any motion state. \\

Zelda \cite{romero2023zelda} & 2023 & LGR & VTM &  Introduce a video analytic system using VLMs that delivers semantically diverse, high-quality results.\\ 

BEV-CLIP \cite{BEV-CLIP} & 2024 & LGR & VTM & Present a BEV traffic scene retrieval method using descriptive text as input.\\

MGNLVR \cite{MGNLVR} & 2022 & LGR & VTM & Introduce a multi-granularity system for natural language-based vehicle retrieval. \\

SSDA-CLIP \cite{le2023trackedVehicleRetrieval} & 2023 & LGR & VTM & Propose a domain-adaptive CLIP model with semi-supervised training for vehicle retrieval. \\

MLVR \cite{xie2023unifiedStructureforRetrieving}& 2023 & LGR & VTM &  Propose Language-guided Vehicle Retrieval system for retrieving the trajectory of tracked vehicles. \\

NIC \cite{Wei2020NIC} & 2020 & IC & V2T, VTM &  Introduce image captioning task to autonomous driving for traffic scene understanding. \\

AnomalyCLIP \cite{anomalyclip} & 2023 & AR & VTM & Propose the first method for VAR based on LLM models to detect and classify anomalous events. \\

LLM-AD \cite{Elhafsi2023LLM-AD} & 2023 & AR & LLM & Propose a framework to detect semantic anomalies leveraging LLMs’ reasoning abilities. \\

NuScenes-QA \cite{qian2023nuscenesQA} & 2023 & VQA & M2T, VTF &   Introduce visual question-answering task and baseline model in autonomous driving based on nuScenes.  \\

NuScenes-MQA \cite{NuScenes-MQA} & 2023 & VQA, IC & M2T& Introduce a Markup-QA dataset based on NuScenes with baseline model.\\

\multirow{2}{*}{Talk2BEV \cite{dewangan2023talk2bev}} & \multirow{2}{*}{2023} & \multirow{2}{*}{VQA, OL-DM} & \multirow{2}{*}{M2T} & Augment BEV maps with language to enable general-purpose visuolinguistic reasoning for autonomous\\
& & & &   driving scenarios. \\

LiDAR-LLM \cite{LiDAR-LLM} & 2023 & VQA, 3D-G-C & LLM & Introduce a LLM with LiDAR data as input and perform 3D scene understanding tasks.  \\

Reason2Drive \cite{Reason2Drive} & 2023 & VQA & V2T, VTF & Propose a method empowering VLMs to leverage object-level perceptual elements.  \\

TRIVIA \cite{qasemi2023traffic} & 2023 & VQA, IC & VTF& Present a novel approach termed Traffic-domain Video Question Answering with Automatic Captioning. \\

CMQR \cite{liu2023causality} & 2023 & VQA & VTM, VTF & Propose a causality-aware event-level visual question reasoning framework to achieve robust VQA. \\

\multirow{2}{*}{Tem-adapter \cite{chen2023tem}} & \multirow{2}{*}{2023} & \multirow{2}{*}{VQA} & \multirow{2}{*}{VTM} & Propose an adapter to enable the learning of temporal dynamics and complex semantics by a visual \\
& & & & temporal aligner and a textual semantic aligner. \\

\midrule
\multicolumn{5}{c}{\textbf{Navigation and Planning}} \\ 
\midrule

TttV \cite{sriram2019talk} & 2019 & LGN & M2T, VTF &  Propose a modular framework to provide accurate waypoints given language instructions.\\  
 
GtN \cite{jain2023groundthennavigate} & 2022 & LGN & M2T, VTM & Present a vision language navigation tool to control vehicle movements based on text-driven commands. \\

ALT-Pilot \cite{omama2023alt} & 2023 & LGN & M2T, VTM & Propose an autonomous navigation system based on language-augmented topometric maps. \\

GPT-Driver \cite{mao2023gpt} & 2023 & MP & LLM & Transform motion planning task into a language modeling problem.\\

CoverNet-T \cite{CovertNet} & 2023 & TP & M2T, VTF & Introduce text and image representation for trajectory prediction. \\

\multirow{2}{*}{DriveVLM \cite{DriveVLM} } & \multirow{2}{*}{2024} & \multirow{2}{*}{TP, MP} & \multirow{2}{*}{M2T, M2F} &  Propose a pipeline incorporating CoT module for hierarchical planning, and a hybrid system enhancing \\ 
& & & & spatial understanding. \\

\midrule
\multicolumn{5}{c}{\textbf{Decision Making and Control}} \\ 
\bottomrule

Advisable-DM \cite{kim2020advisable} & 2020 & OL-DM & M2T, VTF & Propose an advisable and explainable model for self-driving systems.\\

\multirow{2}{*}{LanguageMPC \cite{sha2023languagempc}} & \multirow{2}{*}{2023} & \multirow{2}{*}{OL-DM} & \multirow{2}{*}{LLM} & Devised a chain-of-thought framework for LLMs in driving scenarios that divides decision-making \\
& & & &  process into numerous sub-problems.\\

DaYS \cite{drivingasyouspeak} & 2023 & OL-DM, MP & LLM & Provide a framework to integrate LLMs into autonomous vehicles. \\

\multirow{2}{*}{BEVGPT \cite{BEVGPT} } & \multirow{2}{*}{2023} & \multirow{2}{*}{OL-DM, MP} & \multirow{2}{*}{M2T} &  IPropose a generative pre-trained large language model for scenario prediction, decision-making,  \\ 
& & & & and motion planning with BEV images as input. \\

DwLLMs \cite{Drivingwithllms}& 2023 & OL-C, VSR & LLM & Propose a model with object-level fusion for Explainable AD. \\

DiLU \cite{wen2023dilu} & 2023 & CL-DM & LLM & Instill knowledge-driven capability into AD systems from the perspective of how humans drive. \\

SurrealDriver \cite{jin2023surrealdriver} & 2023 & CL-DM & LLM & Develop an LLM-based driver agent for complex urban environments. \\

DLaH \cite{fu2023drivelikeahuman} & 2023 & CL-DM & M2T, VTF & Show the feasibility and decision-making ability of LLM in driving scenarios in simulation. \\

\bottomrule
\multicolumn{5}{c}{\textbf{End-to-End Autonomous Driving}} \\ 
\bottomrule

\multirow{2}{*}{DriveGPT4 \cite{xu2023drivegpt4}} & \multirow{2}{*}{2023} & \multirow{2}{*}{VQA, OL-C} & \multirow{2}{*}{M2T, VTF} &  Propose a multimodal model for interpretable autonomous driving with video and question as \\ 
& & & & input and control signal as output. \\

ADAPT \cite{jin2023adapt} & 2023 & VSR, OL-DM & V2T, VTF &  Propose a end-to-end action narration and reasoning framework for self-driving vehicles.\\

DriveMLM \cite{DriveMLM} & 2023 & CL-C & M2T, VTF & Propose a LLM-based framework enabling closed-loop AD in simulator.\\

\multirow{2}{*}{VLP \cite{VLP} } & \multirow{2}{*}{2024} & \multirow{2}{*}{OL-P, 3D OD-T} & \multirow{2}{*}{M2T, VTF} &  Introduce a vision language planning approach to enhance the perception and open-loop\\ 
& & & & planning performance. \\

\bottomrule
\multicolumn{5}{c}{\textbf{Data Generation}} \\ 
\bottomrule

DriveGAN \cite{drivegan} & 2021 & CVG & M2V, VTF & Propose an end-to-end controllable differentiable neural driving simulator for scenario re-creation. \\

\multirow{2}{*}{GAIA-1 \cite{Hu2023GAIA-1}} & \multirow{2}{*}{2023} & \multirow{2}{*}{CVG} & \multirow{2}{*}{M2V, VTF} & Propose a multi-modal generative model to produce realistic driving scenarios with precise  \\ 
& & & & control over ego-vehicle actions and scene attributes. \\

\multirow{2}{*}{DriveDreamer \cite{wang2023drivedreamer}} & \multirow{2}{*}{2023} & \multirow{2}{*}{CVG} & \multirow{2}{*}{M2V, VTF} & Introduce the first world model derived from real-world driving scenarios, capable of generating \\
& & & &  high-quality driving videos and sound driving policies. \\

BEVControl\cite{BEVcontrol} & 2023 & CIG & M2V, VTF  & Present a sketch-based street-view images generative model based on nuScenes. \\

DrivingDiffusion\cite{li2023drivingdiffusion} & 2023 & CVG & M2V, VTF  & Propose a spatial-temporal consistent diffusion framework to generate layout-guided multi-view videos.  \\

ADriver-I \cite{ADriver-I} & 2023 & CVG & M2T, M2V & Construct as world model using vision-action pairs based on VLMs. \\

\bottomrule
\end{tabular}
}
\label{vlms_ad}
\end{table*}

\noindent \textbf{Language-Guided Retrieval.} Language-guided object and scene retrieval is a widely applied task in autonomous driving and can be utilized for data selection, corner case retrieval, and others. Zelda \cite{romero2023zelda} employs the semantic similarity capabilities of VLMs for Video Retrieval (VR) and tests its performance in traffic scenes, achieving performance that surpasses other state-of-the-art video analytics systems. BEV-CLIP \cite{BEV-CLIP} proposes a text-to-bird's eye view (BEV) retrieval approach, which utilizes text descriptions to retrieve corresponding scenes with BEV features. It incorporates LLMs as the text encoder to generalize its zero-shot inference ability. The AI City Challenge \cite{6thAICityChallenge}\cite{7thAICityChallenge} has been hosting the Tracked-Vehicle Retrieval by Natural Language Descriptions as a challenge track. As part of this collective effort, \cite{MGNLVR} proposes a multi-granularity retrieval approach for natural language-based vehicle retrieval. The key concept is the introduced multi-query retrieval module based on language augmentation. The idea behind this module is to leverage multiple imperfect language descriptions to achieve higher robustness and accuracy. An interesting approach for solving out-of-distribution input data for vehicle retrieval has been proposed by \cite{le2023trackedVehicleRetrieval}. The key contribution of this work is the introduced domain adaptive training method, which transfers knowledge from labeled data to unseen data by generating pseudo labels. MLVR \cite{xie2023unifiedStructureforRetrieving} proposes a multi-modal language vehicle retrieval framework that employs text and image extractors for feature encoding, subsequently generating video vector sequences through a video recognition module. By integrating modules that combine various vehicle characteristics, MLVR creates more informative vehicle vectors for matching control and accomplishes language-guided retrieval. \\

\noindent \textbf{Driving Scene Understanding.} Precise and high-level driving understanding is critical for ensuring the safety of fully automated driving and building the basis for reasonable decision-making. In Fig.~\ref{fig:vlm_test}, we illustrate examples of the understanding ability of GPT-4V \cite{GPT4V} in a traffic accident scenario and an urban road scene with potential risk. There are several exploratory works utilizing VLMs to understand traffic scenes through specific downstream tasks. \cite{Wei2020NIC} attempts to understand traffic scenes by describing images of the scenes through Image Captioning (IC). The recognition of anomaly scenes in autonomous driving is also of great importance. AnomalyCLIP \cite{anomalyclip} employs the CLIP model for video anomaly detection. By specifying anomaly categories and using context optimization \cite{ContextOptimization}, it distinguishes between normal and abnormal samples, enabling the model to identify anomalous instances. AnomalyCLIP achieves good results in various datasets, including traffic-scene anomaly detection. \cite{Elhafsi2023LLM-AD} transforms visual information into language descriptions and then leverages the strong reasoning capabilities of LLMs to address Semantic Anomaly Detection (SAD).  \\


\noindent \textbf{Visual Scene Reasoning.}
Another emerging field is traffic visual scene event understanding, which commonly forms a Visual Question Answering (VQA) task, as introduced in Section \ref{Traffic Scene Understanding task}. To create a fair evaluation and comparison, NuScenes-QA \cite{qian2023nuscenesQA} proposes a QA set based on the nuScenes dataset and establishes a benchmark for VQA tasks in the autonomous driving scene, providing a foundation for subsequent research. 

Similarly, NuScenes-MQA \cite{NuScenes-MQA} proposes a Markup-QA dataset in which the QAs are enclosed with markups, also based on the nuScenes dataset. Talk2BEV \cite{dewangan2023talk2bev} employs a BEV-based detection model, dense captioning model, and text recognition model to construct a ground-truth language-enhanced BEV map and evaluate the model's performance in visual and spatial understanding based on VQA tasks. LiDAR-LLM \cite{LiDAR-LLM} proposes an LLM that takes raw LiDAR data as input and is able to perform multiple 3D understanding tasks, including 3D captioning, 3D grounding, and 3D VQA. With the proposed View-Aware Transformer, LiDAR-LLM adaptively incorporates LiDAR information as an input feature of LLM and generates the response together with the input prompt. Reason2Drive\cite{Reason2Drive} presents a benchmark for explainable reasoning in complex driving environments. It divides the complex decision-making process into perception, prediction, and reasoning steps and collects question-answer pairs from nuScenes\cite{caesar2020nuscenes}, Waymo\cite{waymo}, and ONCE\cite{once}. It also proposes a framework leveraging object-level information as the baseline. \cite{SUVQATianming} evaluates the VQA performance specifically on road sign and traffic signal existence problems based on BDD100K\cite{yu2020bdd100k}. 

In the traffic domain, \cite{qasemi2023traffic} proposes a weakly supervised Traffic-domain Video Question Answering with Automatic Captioning method. The core contribution is the usage of automatically generated synthetic captions for online available urban traffic videos. The automatically generated video-caption pairs are then used for fine-tuning, thus injecting additional traffic domain knowledge into the trained model. \cite{liu2023causality} proposes a Cross-Modal Question Reasoning framework to identify the temporal causal context for event-level question reasoning. An attention-based module enables the learning of temporal causal scenes and question pairs. \cite{chen2023tem} introduces Tem-Adapter to minimize the gap between image and video domains from the temporal aspect by learning temporal dependencies. It shows great performance in traffic video question-answering tasks. \\

\noindent \textbf{Pedestrian Detection.} Human-like object confusion and insufficient border case samples are the inherent challenges in pedestrian detection, while VTM-based VLMs can generate free and cheap pseudo labels, which can be utilized to alleviate these challenges. To this end, VLPD \cite{liu2023vlpd} first proposes a vision-language extra-annotation-free method to enhance the model's capability of distinguishing confusing human-like objects. It employs CLIP to acquire the pixel-wise explicit semantic contexts and distance pedestrian features from the features of other categories through contrastive learning, improving the detection capabilities for broader cases. UMPD \cite{liu2023umpd} also utilizes the zero-shot semantic classes from the CLIP and proposes a fully-unsupervised multi-view Pedestrian Detection approach without manual annotations. \\

\noindent We note that the aforementioned works, including object referring, open-vocabulary detection, segmentation and tracking, language-guided retrievals, driving scene understanding, and visual scene reasoning in AD, are still in the early stages. Still, we believe these are promising directions and anticipate an increasing number of interesting works to emerge in the future. 

\subsection{Navigation and Planning} 
\label{naivigation_palnning}
In the field of navigation, with the advancement of VLMs, especially the proposal of CLIP \cite{CLIP}, Language-Guided Navigation (LGN) task begins to extend from specific pre-defined location descriptions to free and arbitrary instructions, which also promotes the development of Language-Augmented Maps \cite{omama2023alt}. \\

\noindent \textbf{Language-Guided Navigation.} As formulated in Section \ref{Language-Guided Navigation}, the development of language-guided aviation tasks greatly benefits from the cross-modality alignment in VTM-type VLMs. Talk to the Vehicle \cite{talktothevehicle} proposes a waypoint generator network (WGN) that maps semantic occupancy and pre-defined natural language encodings (NLE) to local waypoints. The planning module then takes local waypoints to predict the trajectory for execution. Ground then Navigate \cite{jain2023groundthennavigate} solves the language-guided navigation tasks with the help of CLIP. It proposes a pipeline that takes video frames, historical trajectory context, and language command as input, and output predicted navigation mask as well as trajectory at each timestamp. ALT-Pilot \cite{omama2023alt} enhances OpenStreetMap (OSM) road networks by incorporating linguistic landmarks, including street signs, traffic signals, and other prominent environmental features that aid in localization to substitute traditional memory and compute expensive HD LiDAR maps. ALT-Pilot also leverages CLIP to precompute the feature descriptors for each of these landmarks and match them with the pixel-wise vision descriptors using cosine similarity at inference time, which facilitates the correspondence from language navigation instructions to map locations, thereby assisting multi-modal localization and navigation. \\

\noindent \textbf{Prediction and Planning.} Some works have also begun to explore how to leverage LLMs to enhance the performance of motion planning and trajectory predictions. GPT-driver \cite{mao2023gpt} reformulates motion planning as a language modeling problem and transforms the GPT-3.5 model into a motion planner for autonomous driving, capitalizing on its strong reasoning and generalization capabilities. CoverNet-T \cite{CovertNet} proposes to train a joint encoder with text-based scene descriptions and rasterized scene images for trajectory predictions. It shows that text-based scene representation has complementary strengths to image encodings, and a joint encoder outperforms its individual counterparts. DriveVLM \cite{DriveVLM} introduces a scene understanding and planning pipeline based on Qwen-VL\cite{bai2023qwenvl} with chain-of-thought (CoT) reasoning. It divides the planning tasks into scene description, scene analysis, and hierarchical planning and calculates the planning results step by step. To mitigate the spatial reasoning and heavy computation limitations of large VLMs, it proposes incorporating traditional autonomous driving modules, including 3D perception, motion prediction, and trajectory planning and formulates a hybrid system.

\subsection{Decision-Making and Control}
\label{Decision-Making and Control}

In the decision-making and control area of autonomous driving, several works seek to leverage the powerful common sense comprehending and reasoning capabilities of LLMs to assist drivers \cite{kim2020advisable}\cite{drivingasyouspeak}, to emulate and entirely substitute drivers \cite{fu2023drivelikeahuman}\cite{sha2023languagempc}\cite{Drivingwithllms, wen2023dilu, jin2023surrealdriver}. When considering utilizing LLMs for closed-loop control in AD, most works \cite{fu2023drivelikeahuman}\cite{drivingasyouspeak}\cite{wen2023dilu}\cite{jin2023surrealdriver} introduce the extra memory module to record driving scenarios, experiences, and other essential driving information. \\

\noindent \textbf{Decision Making.} LanguageMPC \cite{sha2023languagempc} employs LLMs as a decision-making component to solve complex AD scenarios that require human commonsense understanding. Drive as You Speak \cite{drivingasyouspeak} proposes a framework that integrates LLMs into AD and orchestrates other modules accordingly. Drivers can directly communicate with the vehicle through LLMs. The framework includes a memory module to save past driving scenario experiences in a vector database, which includes decision cues, reasoning processes, and other valuable information. LLMs then make decisions based on acquired experience and common sense. DiLU \cite{wen2023dilu} studies the driving methods of human drivers and proposes a paradigm that uses reasoning, memory, and reflection modules to facilitate interaction between LLMs and the environment. This approach embeds such knowledge-driven capabilities of human drivers into the AD system.

BEVGPT\cite{BEVGPT} proposes a framework that integrates several autonomous driving tasks, encompassing driving scenario prediction, decision-making, and motion planning using a single LLM with BEV representation as a unified input. DwLLMs \cite{Drivingwithllms} encodes traffic participants and environment into object-level vectors. It introduces a new paradigm with a two-stage pre-training and fine-tuning approach, allowing the model to understand driving scenarios and generate driving actions. SurrealDriver \cite{jin2023surrealdriver} proposes a human-like AD framework based on LLMs within the CARLA simulator. Through memory and safety mechanisms, LLMs can accomplish situation understanding, decision-making, and action generation. It also learns the driving habits of human drivers and continuously optimizes the driving skills in the closed loop. DLaH \cite{fu2023drivelikeahuman} introduces reasoning, interpretation, and memory modules to construct an AD system based on GPT-3.5\cite{LLMs5GPT3} and LLaMA-Adapter v2\cite{gao2023llamaadapterv2}. It demonstrates strong capabilities in scenario understanding and addressing long-tail problems in simulations. \\

\noindent Although most of the existing control and decision-making work in AD relies solely on LLMs, they can easily get connected with the perception module utilizing visual-LLMs connectors \cite{gao2023llamaadapter,gao2023llamaadapterv2,liu2023visualllava,liu2023improvedllava}, achieving either mid-to-mid or end-to-end AD. Furthermore, designing a vision-LLMs connector suitable for AD systems is a promising direction. We encourage exploration in this area and believe that a substantial amount of work in this segment will emerge in the near future.

\subsection{End-to-End Autonomous Driving}
\label{End-to-End Autonomous Driving}


According to the definition in \cite{End-to-endADSurvy}, the end-to-end autonomous driving system is a fully differentiable program that takes raw sensor data as input and produces a plan or low-level control actions as output, which aligns well with the structure of M2T model in VLMs. Due to such a natural synergy, some studies explore the feasibility of applying M2T VLMs models to end-to-end AD, as illustrated in Figure \ref{fig:3typeVLMs}. Compared to traditional end-to-end AD systems, large VLMs-based end-to-end AD systems possess potent interpretability, trustworthiness, and complex scene comprehending ability, paving the way for the practical application and implementation of end-to-end autonomous driving. \\

\noindent \textbf{End2End AD System.} DriveGPT4 \cite{xu2023drivegpt4} is the pioneering work leveraging large VLMs for end-to-end AD tasks, which takes raw sensor data and human questions as input and outputs predicted control signals and corresponding answers. It retains LLMs' powerful zero-shot generation capability and is able to handle unseen scenarios. 
ADAPT \cite{jin2023adapt} proposes an end-to-end AD pipeline based on the transformer model. With video input, ADAPT continuously outputs control signals, narration, and reasoning descriptions for actions. Unlike the DriveGPT4, ADAPT does not incorporate VQA module but instead transforms interpretable end2end AD into the vision captioning task. DriveMLM \cite{DriveMLM} introduces an LLM-based AD framework, which can plug and play in the existing closed-loop driving systems. The model takes driving rules, user commands, and camera or LiDAR information as input, and generates control signals and corresponding explanations. DriveMLM shows its effectiveness on closed-loop driving in Town05Long benchmark \cite{town05}. VLP \cite{VLP} proposes a vision-language planning framework leveraging  LLMs for end-to-end autonomous driving. The framework incorporates local and global context features into LLMs with the proposed agent-centric and self-driving-car-centric learning paradigms. With the help of LLMs' powerful zero-shot generalization ability, VLP shows improved performance in new urban environments. \\

\noindent The natural fit of M2T-Type VLMs and end-to-end autonomous driving systems has promoted a number of pioneering studies in the field and demonstrates its feasibility and superiority. However, due to the inherent limitations of auto-regressive large VLMs and LLMs, many issues still need to be considered and addressed, and more discussion can be found in Section \ref{Challenges}. 

\begin{figure}[t]
    \centering
   \includegraphics[width=0.49\textwidth]{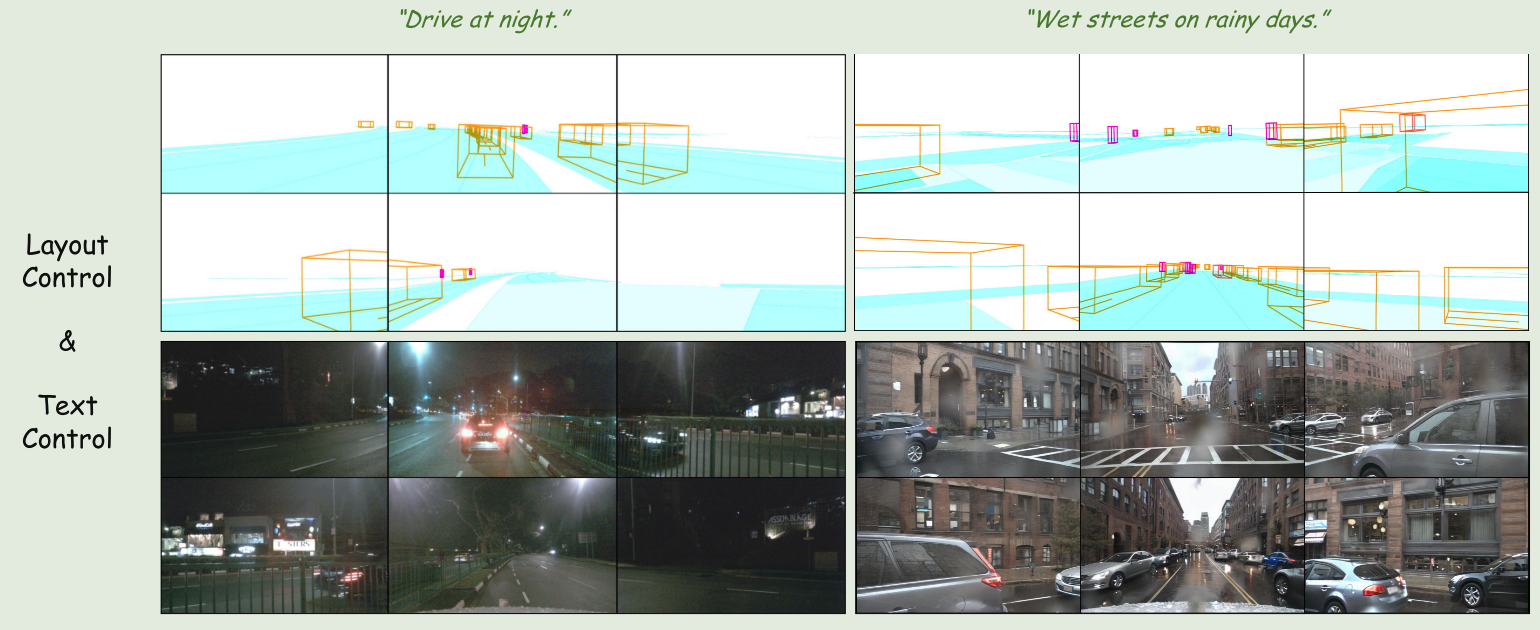}
    \caption{Example of conditional multi-view video generation result from DrivingDiffusion\cite{li2023drivingdiffusion} with layout and text control. }
    \label{fig:drivingdiffusionexample}
\end{figure}

\subsection{Data Generation}
\label{data_generation}
Conditional autonomous driving data generation task, as introduced in Section \ref{data_generation_tasks}, significantly benefits from the advancement and success of generative networks \cite{generativemodel1,generativemodel2,generativemodel3,generativemodel4,generativemodel5,generativemodel6,generativemodel8}. The application of conditional generative models in autonomous driving allows the generation of large-scale high-quality data, thereby promoting the development of data-driven autonomous driving. Table \ref{Data Generation Models Table} compares current mainstream methods and shows the performance in this area. Figure \ref{fig:drivingdiffusionexample} shows an example of generated multi-view frames with different layouts and texts as conditions.

DriveGAN \cite{drivegan} learns the sequences of driving videos and their corresponding control signals. It can control the vehicle behaviors in the generated video by disentangling scene components into action-dependent and action-independent features. This capability enables high-fidelity, controllable neural simulations, and AD data generation. BEVControl \cite{BEVcontrol} takes a sketch-style BEV layout and text prompts as inputs to generate street-view multi-view images. It introduces controller and coordinator elements to ensure the geometrical consistency between the sketch and output and strengthens appearance consistency across the multi-view images. This approach facilitates controllable AD scenario sample generation based on BEV sketch. DrivingDiffusion \cite{li2023drivingdiffusion} generates controllable multi-view videos given 3D layouts and text prompts based on latent diffusion model (LDM)\cite{ldm} and denoising diffusion probabilistic model (DDPM)\cite{ddpm}. To ensure cross-view and cross-frame consistency, a consistency attention mechanism and geometric constraints are introduced to enhance the performance of generated data.  \\

\begin{table}[t]
\centering
\caption{Performance comparison of existing conditioned autonomous driving data generation models on nuScenes validation dataset. Multi-View and Multi-Frame denote the type of generated visual data.}
\renewcommand{\arraystretch}{1.1}
\begin{tabular}{lcccc}
\toprule[1pt]
\textbf{Method} & \textbf{Multi-View} & \textbf{Multi-Frame} & \textbf{FID↓} & \textbf{FVD↓} \\    
\midrule[0.7pt]
DriveGAN \cite{drivegan} & & \checkmark & 73.4 & 502.3\\
BEVGen \cite{BEVGen} & \checkmark & & 25.54 & -    \\
DriveDreamer \cite{wang2023drivedreamer} & & \checkmark &  52.6  & 452.0  \\
DrivingDiffusion \cite{li2023drivingdiffusion} & \checkmark & \checkmark & 15.83 & 332.0  \\ 
ADriver-I \cite{ADriver-I} & & \checkmark & 5.5 & 97.0  \\ 
\bottomrule[1pt]
\end{tabular}
\label{Data Generation Models Table}
\end{table}

\noindent \textbf{World Models.} Some works incorporate the idea of world models \cite{LeCun2022} into AD data generation for a more reasonable, predictable, and structured environment simulation. DriveDreamer \cite{wang2023drivedreamer} is a pioneering world model in AD entirely learned from real-world driving scenarios, which undergoes two stages of training. It first comprehends and models driving scenarios using real-world driving videos, allowing it to acquire detailed and structured environmental information. In the second stage, it constructs a driving world model through a video prediction task, gaining the ability to forecast future events and interact with the environment. DriveDreamer generates realistic and controllable driving scenarios, which can be used for AD model training. GAIA-1 \cite{Hu2023GAIA-1} takes video, action, and text description as inputs, utilizing the powerful capability of world models to learn structured representation and understand the environment, encoding the inputs into a sequence of tokens. It then employs denoising video diffusion models as the video decoder to achieve highly realistic video. ADriver-I \cite{ADriver-I} introduces the interleaved vision-action pair to fuse the visual and control signal. It uses historical vision-action pairs and current images to predict the control signal of the current frame and then generate the next frame image in an auto-regressive manner.

\section{Datasets} 
\label{dataset}

\begin{table*}[htb!]
\centering
\caption{Overview of dataset. \textbf{SS}: Semantic Segmentation \textbf{OT}: Object Tracking \textbf{ReID}: Re-Identification \textbf{MP}: Motion Planning \textbf{AE}: Action Explanation \textbf{VSR}: Visual-Spatial Reasoning \textbf{VLN}: Visual-Language Navigation \textbf{H2V-Advice}: Human-to-Vehicle Advice \textbf{SOR}: Single Object Referring \textbf{VR}: Vehicle Retrieval \textbf{VQA}: Visual Question Answering \textbf{DM}: Decision-Making \textbf{IC}: Image Captioning \textbf{IR}: Importance Ranking }
\resizebox{\textwidth}{!}{
    \begin{tabular}{cccccccc}
    \toprule[1pt]
        \multirow{2}{*}{\textbf{Dataset}} & \multirow{2}{*}{\textbf{Year}} & \multirow{2}{*}{\textbf{Tasks}} & \multirow{2}{*}{\textbf{Source Datasets}} & \multicolumn{4}{c}{\textbf{Data Modalities}} \\
        ~ & ~ & ~ & ~ & Image & Video & Point Cloud & Text \\ \midrule[0.7pt]
        
        \multicolumn{8}{c}{Autonomous Driving Dataset} \\ 
        \midrule[0.7pt]
        Caltech Ped Det \cite{dollar2009pedestrian} & 2009 & 2D OD & - & \checkmark & - & - & - \\
        KITTI \cite{geiger2013vision} & 2012 & 2D/3D OD, SS, OT & - & \checkmark & - & \checkmark & - \\
        Cityscapes \cite{cordts2016cityscapes} & 2016 & 2D/3D OD, SS & - & \checkmark & - & - & - \\
        CityPersons \cite{zhang2017citypersons} & 2017 & 2D OD & - & \checkmark & - & - & - \\
        SemanticKITTI \cite{behley2019semantickitti} & 2019 & 3D SS & - & - & \checkmark & - & - \\
        CityFlow \cite{tang2019cityflow} & 2019 & OT, ReID & - & \checkmark & \checkmark & - & - \\
        nuScenes \cite{caesar2020nuscenes} & 2020 & 2D/3D OD, 2D/3D SS, OT, MP & - & \checkmark & \checkmark & \checkmark & - \\
        BDD100K \cite{yu2020bdd100k} & 2020 & 2D OD, 2D SS, OT & - & \checkmark & \checkmark & - & - \\
        Waymo \cite{waymo} & 2020 & 2D/3D OD, 2D SS, OT & - & \checkmark & \checkmark & \checkmark & - \\
        OTOH \cite{houston2020thousand} & 2020 & MP & - & \checkmark & \checkmark & - & - \\
        ONCE \cite{once} & 2021 & 2D/3D OD & - & \checkmark &\checkmark & \checkmark & - \\
        
        \midrule[0.7pt]

        \multicolumn{8}{c}{Language-Enhanced Autonomous Driving Dataset} \\
        \midrule[0.7pt]
        BDD-X \cite{kim2018textual} & 2018 & AE & BDD100K & \checkmark & \checkmark & - & \checkmark \\
        Cityscapes-Ref \cite{vasudevan2018object} & 2018 & SOR & Cityscapes & \checkmark & - & - & \checkmark \\ 
        TOUCHDOWN \cite{chen2019touchdown} & 2019 & VSR, VLN & - & \checkmark & - & - & \checkmark \\
        LCSD \cite{sriram2019talk} & 2019 & VLN & - & \checkmark & - &- & \checkmark \\
        HAD \cite{Kim2019groundhumantovehicle} &2019 & H2V-Advice& -  & \checkmark & \checkmark &- & \checkmark \\
        Talk2Car \cite{deruyttere2019talk2car} & 2020 & SOR & nuScenes & \checkmark & \checkmark & - & \checkmark \\

        BDD-OIA \cite{bdd-oia} & 2020 & AE & BDD100K & \checkmark & \checkmark & - & \checkmark \\
        
        CityFlow-NL \cite{feng2021cityflow} & 2021 & VR, OT & CityFlow & \checkmark & \checkmark & - & \checkmark \\
        CARLA-NAV \cite{jain2023groundthennavigate} & 2022 & VLN & - & \checkmark & - & - & \checkmark \\
        NuPrompt \cite{wu2023nuprompt} & 2023 & MOR-T & nuScenes & \checkmark & \checkmark & - & \checkmark \\
        NuScenes-QA \cite{qian2023nuscenesQA} & 2023 & VQA & nuScenes & \checkmark & \checkmark & \checkmark & \checkmark \\

        Markup-QA \cite{NuScenes-MQA} & 2023 & VQA & nuScenes & \checkmark & \checkmark & - & \checkmark \\
        Talk2BEV \cite{dewangan2023talk2bev} & 2023 & VQA, OL-DM & nuScenes & \checkmark & \checkmark & - & \checkmark \\

        \multirow{2}{*}{Reason2Drive \cite{Reason2Drive}} & \multirow{2}{*}{2023} & \multirow{2}{*}{VQA} & nuScenes & \multirow{2}{*}{\checkmark} & \multirow{2}{*}{\checkmark} & \multirow{2}{*}{-} & \multirow{2}{*}{\checkmark}  \\
        &&& Waymo,ONCE &&&& \\

        Refer-KITTI \cite{wu2023referrkitti} & 2023 & MOR-T & KITTI & \checkmark & \checkmark & - & \checkmark \\

        Driving LLMs \cite{Drivingwithllms} & 2023 & VSR & - & \checkmark & - & - & \checkmark \\
        DRAMA \cite{malla2023drama} & 2023 & IC, VQA & - & \checkmark & \checkmark & - & \checkmark \\
        Rank2Tell \cite{sachdeva2023rank2tell} & 2023 & IR,VSR & - & \checkmark & \checkmark & \checkmark & \checkmark \\

        SUP-AD \cite{DriveVLM} & 2024 & VQA & - & \checkmark & \checkmark & - & \checkmark \\

        \bottomrule[1pt]
        
    \end{tabular}
}

\label{tab:dataset_tab}
\end{table*}

Open-sourced datasets play a fundamental role in promoting and accelerating the development of autonomous driving areas. Beyond traditional vision-based datasets, integrating the language modality into data offers advantages for driving systems. This section exhibits and analyzes the datasets foundational to autonomous driving (\ref{ad_dataset}) and those integrating language in autonomous driving systems. The overview of datasets is presented in Tab.~\ref{tab:dataset_tab}.

\subsection{Autonomous Driving Datasets} 
\label{ad_dataset}
In the autonomous driving domain, datasets serve as one of the key points for developing safe and efficient perception, prediction, and planning systems. 

Mainstream autonomous driving datasets like KITTI \cite{geiger2013vision}, nuScenes \cite{caesar2020nuscenes}, BDD100K \cite{yu2020bdd100k}, ONCE \cite{once}, and Waymo \cite{waymo} span multiple tasks, such as object detection, tracking, and segmentation, with various data modalities. Cityscapes \cite{cordts2016cityscapes} provides precisely annotated image data for object detection and semantic segmentation. In contrast to the versatile datasets, Caltech Pedestrian Detection \cite{dollar2009pedestrian} offers annotated images for pedestrian detection within urban traffic scenarios. Meanwhile, as a subset of \cite{cordts2016cityscapes}, CityPersons focuses on image-based pedestrian detection from varied city environments. Other task-specific datasets, such as SemanticKITTI \cite{behley2019semantickitti}, provide labeled LiDAR point clouds for semantic segmentation. One thousand and one hours\cite{houston2020thousand} records 1,118 hours of self-driving perception data, which provides trajectory annotations for motion planning tasks. Data given by CityFlow \cite{tang2019cityflow} can be utilized to solve object tracking and re-identification.

\subsection{Language-Enhanced AD Datasets}
\label{vl_dataset_ad}
The introduction of natural language helps the autonomous driving system interpret human questions and instructions, and realize high-level human-vehicle interaction. As autonomous driving advances, combining linguistic information with visual data enriches semantic and contextual comprehension. By facilitating better recognition of the traffic environment and promoting a deeper understanding of driving scenarios, natural language assistance enhances the safety and interaction capabilities of autonomous vehicles.

Prior work \cite{vasudevan2018object} provides a potential opportunity to enhance the capability of the perception system in autonomous vehicles by introducing language understanding into the detector. For object tracking tasks, CityFlow-NL\cite{feng2021cityflow}, Refer-KITTI \cite{wu2023referrkitti}, and NuPrompt \cite{wu2023nuprompt} extend \cite{tang2019cityflow} \cite{geiger2013vision} \cite{caesar2020nuscenes} with language prompts, respectively. TOUCHDOWN \cite{chen2019touchdown}, LCSD \cite{sriram2019talk}, and CARLA-NAV \cite{jain2023groundthennavigate} generate language-guided navigation datasets. Talk2Car \cite{deruyttere2019talk2car} is proposed for the single traffic object referring task. Safe autonomous driving requires reliable scene understanding. \cite{qian2023nuscenesQA}\cite{Drivingwithllms}\cite{malla2023drama}\cite{NuScenes-MQA}\cite{Reason2Drive} evaluate the understanding and reasoning capabilities of the autonomous driving system by providing question-answer pairs. Talk2BEV \cite{dewangan2023talk2bev} focuses on visual-spatial reasoning (VSR). Beyond image and video data, Rank2Tell \cite{sachdeva2023rank2tell} takes LiDAR point clouds into account for multi-modal importance ranking and reasoning. BDD-X \cite{kim2018textual} and BDD-OIA \cite{bdd-oia} offer textual explanations for improving the explainability of AD algorithms. HAD \cite{Kim2019groundhumantovehicle} proposes a human-to-vehicle advice dataset for developing advisable autonomous driving models.

\newpage
\section{Discussion}
\label{discussion}

Based on the aforementioned categorization, analysis, and summary of existing studies, in this section, we deeply discuss benefits, advantages, and several notable emerging topics related to applying vision language models in autonomous driving, including the advantages of large VLMs in AD, Autonomous Driving Foundation Model, Multi-Modality Adaptation, Public Data Availability and Formatting, and Cooperative Driving System with VLMs. \\

\begin{figure}[htb]
    \centering
   \includegraphics[width=0.48\textwidth]{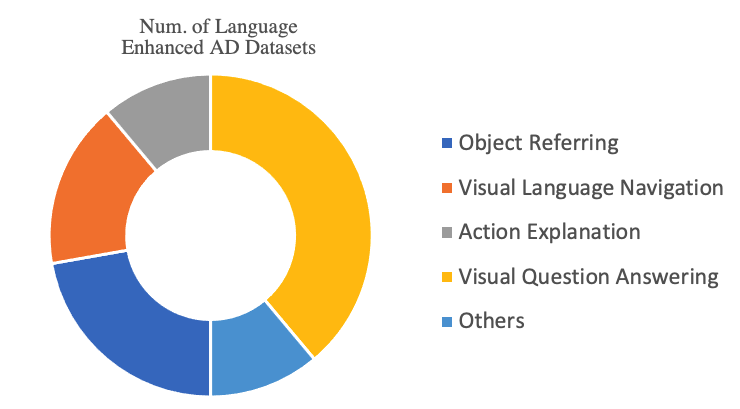}
    \caption{An overview of statistics on the distribution of autonomous driving language-enhanced datasets in table \ref{tab:dataset_tab}. The visual question-answering type has the highest percentage.}
    \label{fig:Num_of_dataset}
\end{figure}

\noindent \textbf{Advantages of large VLMs in AD.} 
Autonomous driving has long faced a series of challenges, including issues with action explainability, high-level decision-making, generalization to complex and extreme scenarios, and human-vehicle interaction. The emergence of LLMs and large VLMs offers new potential solutions to all these problems. Large VLMs, in particular, provide comprehensive benefits for autonomous driving, enhancing and expanding the capabilities of classic AD modules, such as perception, navigation, planning, and control, and also offer a fresh perspective on the end-to-end autonomous driving paradigm. 

Specifically, as mentioned in \ref{perception_understanding}, large VLMs leverage the outstanding zero-shot generalization capabilities to enhance the recognition abilities in corner cases and unknown objects. By harnessing the deep fusion of scene and textual information, large VLMs provide plausible open-ended QA sessions regarding driving-related scenarios, facilitating explainable autonomous driving and human-vehicle interaction. Furthermore, as stated in section \ref{naivigation_palnning} and \ref{End-to-End Autonomous Driving}, large VLMs promote the performance of language-guided localizing, navigation, and planning tasks with the ability to align cross-modality features and comprehend in-depth instructions. Its advanced high-level reasoning abilities, along with an understanding of common sense and traffic rules, enable the potential to perform high-level decision-making and control tasks according to section \ref{Decision-Making and Control}. Lastly, as noted in section \ref{data_generation}, large VLMs facilitate the generation of realistic, controllable, high-quality synthetic autonomous driving data with comprehensive and implicit modeling of the environment. In summary, large VLMs have introduced new perspectives and opportunities for autonomous driving. Existing studies have conducted feasibility verification in various aspects, and many areas are worth further exploration. \\

\noindent \textbf{Autonomous Driving Foundation Model.} 
Existing foundation models—including vision foundation models \cite{wang2023seggpt,kirillov2023segmentanything,wang2023painter}, language foundation models \cite{LLMs1llama2,LLMs2LLama,LLMs3Bloom}, and multi-modal foundation models \cite{zhu2023minigpt,peng2023kosmos,gong2023multimodal}—have set the stage for the feasibility of Autonomous Driving Foundation Models (ADFMs). We formulate ADFMs as models pre-trained on vast and diverse datasets, excelling in interpretability, reasoning, forecasting, and introspection, and effective in various autonomous driving tasks, such as perception, understanding, planning, control, and decision-making. Some studies have made preliminary attempts \cite{xu2023drivegpt4}\cite{Hu2023GAIA-1}\cite{mao2023gpt}\cite{sha2023languagempc}, whereas how to adapt existing foundation models to ADFMs, and align the objectiveness of autonomous driving remains a relatively uncharted domain. \\

\noindent \textbf{Multi-Modality Adaptation.} 
As mentioned in \ref{Decision-Making and Control}, current approaches utilizing LLMs for motion planning, control, and decision-making often transform sensor data into text formulation directly, either through existing perception algorithms or by direct extraction from the simulator. While this modular approach simplifies experiments, it can lead to the loss of context and environmental information and is heavily dependent on the performance of the perception algorithms. In light of this, exploring the establishment of vision-language connections through VTM or VTF or a hybrid of both, specifically for autonomous driving scenarios, as alternatives to simplistic manual reformulations, is a direction worth pursuing. Some available modality connectors, such as LLAMA-Aapter \cite{gao2023llamaadapter,gao2023llamaadapterv2} and Q-Former\cite{li2023blip2}, can be used as preliminaries. \\

\noindent \textbf{Public Data Availability and Formatting.}    
Although there are already many off-the-shelf large-scale autonomous driving datasets \cite{caesar2020nuscenes}\cite{waymo} available, as mentioned in Section \ref{dataset}, they are not suitable or optimal for the direct adaptation of LLMs in AD. For example, how to generate instruction tuning datasets and design instruction formats based on AD datasets for ADFM adaptations remains barely investigated. Besides, a large-scale image-text traffic-specific pairs dataset can also be of great help for the development of AD, especially for approaches relying on VTM pre-training models in object detection, semantic segmentation, language-guided navigation, and language-guided retrieval.\\

\noindent \textbf{Cooperative Driving System with VLMs.}  
Recent studies \cite{opencda,v2x-vit,Xu_2023_CVPR} demonstrate the advantages and prospects of the cooperative driving system (CDS) as a promising technique for next-generation automated vehicles. Cooperative driving system shares information through vehicle-to-vehicle (V2V) \cite{v2v-cooperative,OPV2V} or vehicle-to-intelligent infrastructure (V2I) \cite{zimmer2024tumtrafv2x} communication to enlarge the perceptive range, alert potential hazards, and optimize the traffic flow. As mentioned in Section \ref{End-to-End Autonomous Driving}, the M2T-type VLM is a natural match for end-to-end autonomous driving, and the vehicle then serves as an intelligent autonomous agent. In this case, CDS can be regarded as a VLM-based collaborative multi-agent system.  

Integrating large VLMs into CDS brings benefits in terms of cooperative perception, cooperative decision-making, and user experience. With the zero-shot traffic environmental perception and understanding ability of large VLMs, CDS can recognize open-vocabulary unseen objects and understand novel corner cases, which further enhances the cooperative perception capability. Besides, CDS can leverage the high-level reasoning ability of large VLMs for cooperative multi-vehicle planning and decision-making. By communicating with other vehicles and infrastructures, all the VLM-based intelligent vehicles can theoretically reach a better planning solution. In addition, with the advantage of large VLMs in interpretability and human-vehicle interaction, leveraging large VLMs in CDS also improves the user traveling experience. The passenger can understand the behavior of other vehicles and the surrounding traffic environment more intuitively and clearly.

\section{Challenges}
\label{Challenges}

Despite the many benefits that large vision language models bring to autonomous driving and their potential to alleviate current problems, there are still some challenges in applying the technology effectively to real-world autonomous driving systems. In this section, we outline the remaining and potential future challenges, present some feasible solutions, and provide insights for further research. \\

\noindent \textbf{Computation Demands and Deployment Latency.}
Real-time processing and limited computational resources pose significant challenges for on-vehicle model deployment in autonomous driving. Current large VLMs typically contain billion-scale parameters, making both fine-tuning and inference highly resource-intensive and failing to meet real-time requirements. Several existing techniques can alleviate these problems. Parameter-Efficient Fine-Tuning (PEFT) \cite{hu2021lora,lester2021powerefficientpt,zhang2023adaptivepeft} reduces the number of trainable parameters while maintaining satisfactory model performance, thereby minimizing resource consumption for fine-tuning. Besides, unlike general LLMs, the knowledge required for autonomous driving is often specialized and domain-specific, and much of the knowledge contained within LLMs is redundant for AD. Hence, employing knowledge distillation \cite{gu2023knowledge}\cite{liu-etal-2022-multi-granularity} to train a smaller, more tailored model suitable for autonomous driving presents a feasible approach. Other common model compression techniques in deep learning, e.g., quantization \cite{xiao2023smoothquant}\cite{bai2022towardsquant} and pruning, are also applicable in this context. 

In addition, it is important to consider the deployment architecture of VLMs in automated vehicles for real-world applicability to meet the computation and latency requirements. On-vehicle deployment benefits from low latency and reduces the risk of privacy leakage with local data storage, but it requires more computational capacity and costs more to produce, update, and maintain. Cloud deployments provide powerful centralized computing capabilities but carry risks of latency, privacy, and network dependency. In fact, a hybrid deployment of large VLMs in AD, leveraging the advantages of the vehicle side and the cloud side separately, is an ideal solution. Critical real-time decisions can be deployed on the vehicle side, whereas data analytics and model updates, which require more computational resources and have fewer speed requirements, can be deployed on the cloud. Currently, both the software architecture and the hardware requirements for the application of large VLMs to automated vehicles are still open questions. \\

\noindent \textbf{Temporal Scene Understanding.}
Scene understanding in autonomous driving typically requires temporal information from video to continuously perceive and comprehend the dynamics and causality of the driving environment and traffic participants. Merely using image-level VLMs is insufficient for the demands. For instance, it's impossible to determine the specific cause of a car accident from a single image; see Fig.~\ref{fig:vlm_test}. However, how to process temporal sensor data for driving scenarios remains an issue that requires exploration. Based on the existing work, one possible direction is to train a video-language model, e.g., \cite{xu2023drivegpt4} as an initial attempt, with all existing video-language adapters \cite{zhang2023videollama}\cite{maaz2023videochatgpt} being potentially applicable in this regard. Another possible route involves converting video data into a paradigm that can be processed by image-language models \cite{chen2023tem}, integrating temporal adapter layers, and fine-tuning accordingly by necessity, thereby enhancing the model's understanding of spatial-temporal information in the traffic environment. \\

\noindent \textbf{Ethical and Societal Concern in Driving Safety.}
Since large language models are usually pre-trained with large amounts of unfiltered, low-cost data, they can sometimes generate toxic, biased, harmful content that may conflict with human values. Besides, the hallucination phenomenon of auto-regressive large language models \cite{huang2023Hallucinationsurvey} also poses serious challenges to practical applications. It brings potential societal concerns regarding model safety and reliability, which necessitates further alignment tuning. Hence, during the development of autonomous driving foundation models, it's also essential to align the control strategy, decision-making, and response mechanisms with certain standards to ensure adherence to stable, safe, and sound driving values. In addition, some ethical issues also arise, such as the attribution of responsibility in the event of traffic accidents, which requires clear delineation in terms of policy. As potential solutions towards safe and trustworthy driving, existing techniques in LLMs alignment tuning, such as reinforcement learning from human feedback (RLHF) \cite{christiano2023deepRLfromHP}, DPO \cite{dpo}, and supervised alignment tuning, can be used to align the controlling behavior and strategy to a reliable driver. These techniques are worth trying in this domain.  \\

\noindent \textbf{Data Privacy and Security.} 
Training large vision language models for autonomous driving not only requires large-scale data from the internet but also some sensitive data that may involve user information, such as user age, driving behavior, road conditions, etc. As a data-driven technology, this valuable data is vital in the journey of achieving fully automated driving. However, how, which, and where this sensitive data can be collected, used, and stored brings many concerns and necessitates further discussion. More regulations should be in place to protect users' privacy, while boosting the development of this field in the meantime. Technically, privacy-preserving machine learning \cite{ppml1,ppml2} and federated learning \cite{fedsurvey1,fedsurvey2} can be possible solutions to alleviate this issue. In addition, the adoption of large VLMs as the core decision-making and planning system in autonomous driving brings a series of security concerns. Security attacks against vehicle system vulnerabilities or adversarial attacks \cite{Adversarialattack} on VLM models can lead to vehicle system manipulation or decision misjudgment. To this end, more research on vehicle adversarial attacks and the security of vehicle system architecture with large VLMs is highly essential for further investigation. \\

\section{Conclusion}
\label{conclusion}

This survey provides a comprehensive overview of the background, mainstream visual language tasks and metrics, current advancements, potential applications, main challenges, and future trajectories for large VLMs in autonomous driving. The paper classifies the existing studies in this domain into Multimodal-to-Text, Multimodal-to-Vision, and Vision-to-Text based on VLM's modality types, and into Vision-Text-Fusion and Vision-Text-Matching from the viewpoint of VLMs' inter-modality connection. It provides exhaustive reviews and analyses of existing VLM applications in various areas of autonomous driving, including perception and understanding, navigation and planning, decision-making and control, end-to-end autonomous driving, and data generation. It summarizes notable datasets available in this field up to the present date and illustrates the advantages of introducing large VLMs into the field. Drawing upon current studies, this work expounds on existing and future challenges encompassing computation demands and latency, temporal scene understanding, ethical and societal concerns in driving safety, and data privacy and security. It also offers a view of some potential solutions and possible directions for future exploration. The aim is to draw interest and attention within the research community to this area and facilitate more meaningful investigations.

{
    \bibliographystyle{ieeetr}
    \bibliography{main}
}

\vspace{5pt}

\vfill

\end{document}